\documentclass[conference]{IEEEtran}
\IEEEoverridecommandlockouts
\usepackage{cite}
\usepackage{amsmath,amssymb,amsfonts}
\usepackage{mathrsfs}
\usepackage[ruled,linesnumbered, vlined]{algorithm2e}
\usepackage{algorithmic}
\usepackage{graphicx}
\usepackage{textcomp}
\usepackage{xcolor}
\usepackage{caption}
\usepackage{multirow}
\usepackage{makecell}
\usepackage{url}
\usepackage{bm}
\usepackage{enumitem}
\usepackage{graphicx}
\usepackage{float}
\usepackage{bbding}
\usepackage{url}
\usepackage{float}
\usepackage{subfig}

\usepackage[normalem]{ulem}

\def\BibTeX{{\rm B\kern-.05em{\sc i\kern-.025em b}\kern-.08em
    T\kern-.1667em\lower.7ex\hbox{E}\kern-.125emX}}
\begin{document}

\title{Search to Fine-tune Pre-trained Graph Neural Networks
	for Graph-level Tasks
}

\author{\IEEEauthorblockN{
		Zhili WANG$^1$,
		Shimin DI$^{1\dag}$\thanks{$^\dag$Corresponding Author}, 
		Lei CHEN$^{1,2}$, 
		Xiaofang ZHOU$^{1}$}
	$^1$The Hong Kong University of Science and Technology, Hong Kong SAR, China\\
	$^2$The Hong Kong University of Science and Technology (Guangzhou), Guangzhou, China \\
	zwangeo@connect.ust.hk, sdiaa@connect.ust.hk,\ leichen@hkust-gz.edu.cn, zxf@cse.ust.hk}

%\and
%\and
%\and

\maketitle

\begin{abstract}
Recently, graph neural networks (GNNs) have shown its unprecedented success in many graph-related tasks.
However, GNNs face the label scarcity issue as other neural networks do.
Thus, recent efforts try to pre-train GNNs on a large-scale unlabeled graph and adapt the knowledge from the unlabeled graph to the target downstream task.
The adaptation is generally achieved by fine-tuning the pre-trained GNNs with a limited number of labeled data.
%{\color{blue}
However, current GNNs pre-training works focus more on how to better pre-train a GNN, but ignore the importance of fine-tuning to better leverage the transferred knowledge.
%}
%Despite the importance of fine-tuning, current GNNs pre-training works often ignore designing a good fine-tuning strategy to better leverage transferred knowledge and improve the performance on downstream tasks.
Only a few works start to investigate a better fine-tuning strategy for pre-trained GNNs.
But their designs either have strong assumptions or overlook the data-aware issue behind various downstream domains.
%{\color{blue}
%Unfortunately, it is a non-trivial task to tackle these issues because
%\textcolor{red}{++two technical challenges++}.
%{\color{teal}
%%First is the 
%%fine-tuning pre-trained GNNs is still a under-investigated problem and 
%there lack a systematic design in existing literatures and it is unclear how to design a powerful space for fine-tuning. 
%Secondly, the entire space equals to the Cartesian product of the fine-tuning strategy space and GNN model parameter space.
%The computational overhead of discovering suitable fine-tuning strategies from such a large space is enormous. 
%}
%}
%{\color{blue}
To further boost pre-trained GNNs,
%} 
we propose to search to fine-tune pre-trained GNNs for graph-level tasks (S2PGNN), which can adaptively design a suitable fine-tuning framework for the given pre-trained GNN and downstream data.
%{\color{blue}
Unfortunately, it is a non-trivial task to 
achieve this goal due to 
two technical challenges.
First is the hardness of fine-tuning space design since 
there lack a systematic and unified exploration in existing literature.
Second is the 
enormous computational overhead required for discovering suitable fine-tuning strategies from the discrete space.
To tackle these challenges,
S2PGNN first carefully summarizes a search space of fine-tuning strategies that is suitable for GNNs, which is expressive enough to enable powerful strategies to be searched.
Then, S2PGNN integrates an efficient search algorithm to solve the computationally expensive search problem from a discrete and large space.
%}
%To ensure the improvement brought by searching fine-tuning strategy, we carefully summarize a proper search space of fine-tuning framework that is suitable for GNNs.
The empirical studies show that S2PGNN can be implemented on the top of 10 famous pre-trained GNNs and consistently improve their performance by 9\% to 17\%.
%Besides, S2PGNN achieves better performance than existing fine-tuning strategies within and outside the GNN area.
Our code is publicly available at 
\url{https://github.com/zwangeo/icde2024}.
%{\color{blue} \url{https://anonymous.4open.science/r/code_icde2024-A9CB/}}.
%\footnote{
%	\textbf{solved} R5O5: The link of the code included in the paper is expired. 
%}
\end{abstract}


%\footnote{\# zhili: {\color{teal} teal color}: revised by zhili}
%\footnote{\# zhili: {\color{brown} brown color}: to be revise/adjust by zhili}
%\footnote{\# zhili: intro to be revised later after related work and method are roughly settled.}

\section{Introduction}
\label{sec: intro}

As one of the most ubiquitous data structures, graph is a powerful way to represent diverse and complex real-world systems, e.g., social networks \cite{guo2020deep}, 
knowledge graphs \cite{shimin2021efficient, di2023message, di2021searching},
protein interactions \cite{tsubaki2019compound}, and molecules \cite{xiong2019pushing, sun2020graph, lu2019molecular,li2022black,li2023early}. Graph representation learning \cite{chen2020graph} maps the original graph into the low-dimensional vector space to handle various graph scenarios.
Recently, Graph Neural Networks (GNNs) \cite{kipf2016semi, hamilton2017inductive, velivckovic2017graph, xu2018powerful, xu2018representation, li2020distance, chien2020adaptive, zhu2020beyond}, which follow the message-passing schema \cite{gilmer2017neural} to learn representations via iteratively neighboring message aggregation, have become the leading approaches towards powerful graph representation learning. GNNs have demonstrated state-of-the-art results in a variety of graph tasks. e.g., node classification \cite{kipf2016semi, hamilton2017inductive, velivckovic2017graph, wang2021autogel, wang2023message}, link prediction \cite{zhang2018link, zhang2019inductive, zhang2020revisiting, li2020distance, wang2021autogel, li2023zebra}, and graph classification \cite{gilmer2017neural, xu2018powerful, ying2018hierarchical, xiong2019pushing, sun2020graph, lu2019molecular, wang2021autogel, wang2023message, wei2023search}. 
%Most GNNs follow the message-passing schema \cite{}, which learn representations via iteratively aggregating messages from neighboring nodes and edges. The specific message-passing schema used in a GNN can vary depending on the data and task at hand \cite{}.

Despite the revolutionary success of GNNs on graph data, they are mainly trained in an end-to-end manner with task-specific supervision, which generally requires abundant labeled data.
However, high-quality and task-specific labels can be scarce, 
which seriously impedes the application of GNNs on 
various graph data~\cite{hu2019strategies}.
%{\color{blue}
Especially, some scientific fields require extensive and laborious expert knowledge for adequate annotation, e.g., medicine, chemistry, and biology.
%} 
Therefore, recent efforts 
%{\color{blue}
\cite{hu2019strategies, you2020graph, xu2021self, zhang2021motif, liu2021pre, xia2022simgrace, hou2022graphmae, xia2022pre}
%}
%\footnote{
%	\textbf{solved} R7-O2: related reference added
%}
investigate pre-training \cite{erhan2010does} in GNNs 
so as to tackle this challenge 
and
improve the generalization performances of GNNs.
%{\color{blue}
Pre-trained GNNs have
demonstrated superiority and
improved generalization performances on the downstream graph-level tasks, e.g., molecular property prediction \cite{hu2019strategies}.
%}
They mainly 
follow the self-supervised way to 
pre-train GNNs on large-scale unlabeled graph data
by 
exploiting
various 
self-supervised learning (SSL) \cite{liu2022graph} 
strategies,
such as 
Autoregressive Modeling (AM) \cite{hu2020gpt, zhang2021motif}, Masked Component Modeling (MCM) \cite{hu2019strategies, xia2022mole}, and Contrastive Learning (CL) \cite{you2020graph, qiu2020gcc, xia2022simgrace, xu2021self}.
%{\color{blue}
Then, due to domain discrepancy \cite{han2021adaptive}, the \textit{fine-tuning} strategy \cite{radenovic2018fine, houlsby2019parameter} is proposed to transfer knowledge from pre-trained GNNs to the downstream domain by training the model with a limited number of labeled data.
%}

Compared with the various pre-training mechanisms, the fine-tuning strategy on pre-trained GNNs has received little attention.
The most common strategy is still the vanilla fine-tuning method \cite{girshick2014rich}, 
where the downstream GNN will be initialized by the parameters of a pre-trained GNN and trained on the labeled data of the targeted domain.
%where the GNN architecture remains unchanged and its parameters are jointly fine-tuned with the additional prediction head. 
%directly inherent the GNN architecture from the 
%keeps the GNN architecture unchanged and jointly fine-tune 
%the model with an additional prediction head are jointly fine-tuned. 
%GNN architecture remains unchanged between pre-training and fine-tuning, and the .
However,
the vanilla strategy may suffer from the issues of
overfitting and poor generalization \cite{xia2022towards, jiang2019smart}.
%{\color{brown}
%inadequate and problematic.
%Generally,
%the success 
%of vanilla fine-tuning heavily
%relies on the 
%consistency 
%between the pre-training and downstream 
%data structures and properties \cite{hu2019strategies, sun2021mocl}, which however may not always hold. 
%}
%For example, in the context of molecular graphs, downstream datasets  usually encompass novel substructures (a.k.a., \textit{scaffold}~\cite{ramsundar2019deep}),
%which have not been encountered during pre-training. 
%Besides,
%downstream graph scenarios in real world are often complex and diverse.
%Blindly utilizing the fixed vanilla strategy to 
%%transfer 
%fine-tune pre-trained GNNs under various downstream scenarios may be inflexible and insufficient.
Few recent efforts \cite{han2021adaptive, xia2022towards, zhang2022fine} dedicate to designing novel 
GNN fine-tuning strategies to 
mitigate potential issues of the 
vanilla solution.
%{\color{blue}
Unfortunately, their designs either have strong assumptions or overlook the data-aware issue behind the various downstream datasets.
Firstly, the strategy~\cite{han2021adaptive} needs the pre-training data and task as prerequisites for the downstream domain, which unfortunately, are often inaccessible.
Others~\cite{xia2022towards, zhang2022fine} assume the high relevance between the pre-trained domain with the downstream one, then enforce the similarity of model parameters or learned representations, which are incapable of handling out-of-distribution predictions.
For example, molecular property prediction \cite{hu2019strategies} often encompass novel substructures that have not been encountered during pre-training, a.k.a., \textit{scaffold}~\cite{ramsundar2019deep}.
Secondly, existing methods follow a fixed strategy to design fine-tuning methods, and their operators and structures remain unchanged when facing different data (i.e., not data-aware).
However, complex and diverse downstream graph data may require data-specific strategy to better conduct generalization.
For example, 
the GNN layers required for different graph data 
may be highly data-specific~\cite{liu2020towards, chien2020adaptive} in terms of density and topology.
Thus, how to selectively and comprehensively fuse
the multi-scale information from different layers of pre-trained GNNs may be a crutial operator towards more data-aware and effective downstream generalization.

To fully unleash the potential of pre-trained GNNs on various downstream datasets, 
we propose a novel idea to search a suitable fine-tuning strategy for the 
given pre-trained GNN and downstream graph dataset.
%namely
%searching to fine-tune pre-trained graph neural networks for graph-level tasks (S2PGNN).
%{\color{blue}
However, it is a non-trivial task to achieve this goal due to two technical challenges. 
Firstly, 
fine-tuning pre-trained GNNs is still a under-investigated problem and there lack a systematic and unified design space in existing literature.
Thus, it is unclear how to design a powerful space for GNN fine-tuning. 
Secondly, the entire space equals to the Cartesian product of the fine-tuning strategy space and GNN model parameter space.
The computational overhead of discovering suitable fine-tuning strategies from such a large space is enormous. 
To address these challenges,
we present 
a novel framework in this paper, named
\textit{S}earch to fine-tune \textit{P}re-trained \textit{G}raph \textit{N}eural \textit{N}etworks for graph-level tasks (S2PGNN).
%We propose to search suitable and specific fine-tuning techniques for given pre-trained GNNs and downstream graph data, 
%which is effective and data-aware.
More concretely, 
we investigate existing literature within and outside the area of GNNs, and systematically summarize a search space of fine-tuning frameworks that is suitable for pre-trained GNNs.
The proposed space includes multiple influential design dimensions and enables powerful fine-tuning strategies to be searched.
To reduce the search cost from the large and discrete space,
we incorporate an efficient search algorithm, which suggests the parameter-sharing and continuous relaxation on the discrete space and solves the search problem by differentiable optimization. 
In summary, the main contributions of this work are listed as follows:
%}
%Specifically, we first 
%investigate the existing GNN fine-tuning works 
%and carefully
%summarize a search space of fine-tuning 
%frameworks that is suitable for GNNs,
%which incorporate influential design dimensions
%and powerful candidates of the existing fine-tuning works.
%Then, we
%leverage an efficient search algorithm to 
%search a concrete fine-tuning strategy from the proposed search space.
%To fully explore and utilize useful designs, we incorporate influential design dimensions and powerful candidates into a unified search space and formulate a search framework, which enables the search of optimal GNN fine-tuning strategy for the given downstream graph data. 
%\footnote{\color{red}
%	Meta-1: Novelty of Contributions: It would be useful for the authors to further highlight and clarify the novelty and technical contributions of the proposed strategy compared with existing approaches, especially the key differences and challenges in designing fine-tuning strategies with pre-trained GNNs for graph-level tasks.
%	
%	R7O3:  It would be better to highlight the key differences and challenges in designing fine-tuning strategies with pre-trained GNNs for graph-level tasks, when compared to node-level tasks.
%}

\begin{itemize}[leftmargin=*]
	
	\item 
	In this paper, we 
	systematically exploring GNN fine-tuning strategies, an important yet seriously under-investigated problem, to improve the utilization of pre-trained GNNs.
	We investigate fine-tuning within and outside GNN area and provide a new perspective for GNN fine-tuning.

	\item 
	To further improve pre-trained GNNs,
	we propose 
	S2PGNN that automatically search a suitable fine-tuning strategy for the given pre-trained GNN
	and downstream graph dataset,
	which broadens the perspective of GNN fine-tuning works.
%	{\color{blue}
	To the best of our knowledge, we are the first to develop automated fine-tuning search framework for pre-trained GNNs.
%	}
	
	\item 
	We propose a novel search space of fine-tuning strategies in S2PGNN,
	which 
	identifies
	key factors that affect GNN fine-tuning results 
	and presents improved strategies.
	The S2PGNN 
	framework
	is model-agnostic and can be plugged into existing
	pre-trained GNNs for better performance.

	\item 
	The empirical studies demonstrate that 
	S2PGNN can be implemented on the top of 10 famous pre-trained GNNs
	and consistently improve their performance.
	Besides, 
	S2PGNN achieves better performance than existing 
	fine-tuning strategies within and outside the GNN area.
	
\end{itemize}

\begin{table}[!t]
	\centering
%	\color{blue}
	\caption{A summary of common notations.}
	\label{tab:notation}
	\begin{tabular}{l|p{4.75cm}}
		\hline
		\textbf{Notation} & \textbf{Definition} \\ 
		\hline
		$\mathbb{R}^{d}$ & The $d$-dimension real space. \\ 
		\hline
%		$G\!=\!(V,E,\mathbf{A},\mathbf{X}^{V}, \mathbf{X}^{E})$ & An attributed graph with node-set $V$, edge-set $E$, adjacency matrix $\mathbf{A}$, and attribute matrices $\mathbf{X}^{V}$ and $\mathbf{X}^{E}$. \\
%		\hline
		$G = (V,E,\mathbf{A})$ & An attributed graph with node-set $V$, edge-set $E$, and adjacency matrix $\mathbf{A}$. \\
		\hline
		$V$, $E$ & \makecell[l]{$V=\{v_1, \dots, v_n\}$, \\$E=\{(v_i, v_j)| v_i, v_j \in V\}$.} \\
		\hline
%		$\mathbf{X}^{V}\in\mathbb{R}^{|V|\times d}, \mathbf{X}_{v} \in \mathbb{R}^{d}$ & Node attribute matrix and vector.\\
%		\hline
%		$\mathbf{X}^{E}\in\mathbb{R}^{|E|\times d}, \mathbf{X}_{uv} \in \mathbb{R}^{d}$ & Edge attribute matrix and vector.\\
%		\hline 
		$\mathbf{X}_{v}, \mathbf{X}_{uv} \in \mathbb{R}^{d}$ & The node and edge attribute vector.\\
		\hline 
		$k$, $K$ & The current and maximum GNN layer index, $1\leq k \leq K$. 
		\\
		\hline 
		$\mathbf{H}_{v}, \mathbf{H}_{G} \in \mathbb{R}^{d}$ 
		& The node and graph representation vector.\\
		\hline 
%		$\mathbf{H}_{G}\!\in\!\mathbb{R}^{d}$ 
%		& The graph representation.\\
%		\hline 
%		$\cdot$& Vector dot product.\\
%		\hline
%		$MLP(\cdot)$
%		& The multi-layer perceptron. \\
%		\hline
%		$||$& Concatenation operator.\\
%		\hline
%		$\sigma (\cdot)$ & Non-linear activation function. \\
%		\hline
%		$tanh (\cdot)$ & Hyperbolic tangent activation function. \\
%		\hline 
		$f_{\bm{\psi}, \bm{\theta}}(\cdot)$ & The GNN encoder with architectures $\bm{\psi}$ and parameters $\bm{\theta}$. \\
		\hline 
%		$g_{\bm{\omega}}(\cdot)$ & The prediction head with parameters $\bm{\omega}$.
%		\\
%		\hline 
		$p_{\bm{\alpha}}(\cdot)$ & The GNN fine-tuning controller with parameters $\bm{\alpha}$. \\
		\hline 
		$\bm{\Phi}_{ft}$ & The GNN fine-tuning strategy. \\
		\hline
%		$f_{aug}(\cdot)$ & The identity augmentation function. \\
%		\hline 
%		$f_{fuse}(\cdot)$ & The multi-scale fusion function. \\
%		\hline 
%		$f_{read}(\cdot)$ & The graph-level readout function. \\
%		\hline
		$f_{dim}(\cdot)$
%		, $f_{id}(\cdot)$, $f_{fuse}(\cdot)$, $f_{read}(\cdot)$
		&The GNN fine-tuning dimension for $dim \in \{conv, id, fuse, read\}$
%		of backbone convolution, identity augmentation, multi-scale fusion, graph-level readout. 
		\\ 
		\hline
		$\mathcal{O}_{dim}$
%		, $\mathcal{O}_{id}$, $\mathcal{O}_{fuse}$, $\mathcal{O}_{read}$
		&\makecell[l]{The candidate-set of $f_{dim}(\cdot)$}\\
%			, $f_{aug}(\cdot)$, $f_{fuse}(\cdot)$, $f_{read}(\cdot)$.} \\ 
		\hline
		$\mathcal{D}_{ssl}$, $\mathcal{D}_{ft}$
		&The pre-training and downstream dataset. \\ 
		\hline
		$\mathcal{L}_{ssl}(\cdot)$, $\mathcal{L}_{ft}(\cdot)$
		&The pre-training and downstream loss. \\ 
		\hline
	\end{tabular}
\end{table}

%\begin{figure}[!t]
%	\centering
%	%	\includegraphics[width=8.25cm]
%	\includegraphics[width=8.75cm]
%	{figures/gnn_pf.png}
%	\caption{The illustration of overall GNN P\&F framework.}
%	\vspace{-10px}
%	\label{fig: gnn_pf}
%\end{figure}

\section{Related Work}
\label{sec: related}

A graph typically can be represented as 
%\sout{$G\!=\!(V, E,\mathbf{A},\mathbf{X}^{V}, \mathbf{X}^{E})$}
$G = (V, E,\mathbf{A})$, 
where $V=\{v_1, \dots, v_n\}$ is 
the node-set,
$E=\{(v_i, v_j)| v_i, v_j \in V\}$ 
is the edge-set, and
$\mathbf{A} \in \{0, 1\}^{|V| \times |V|}$ is the adjacency matrix to define graph topology where $\mathbf{A}_{ij}=1$ iff $(v_i, v_j) \in E$.
%{\color{red}
%\sout{
%$\mathbf{X}^{V}=\{\mathbf{X}_{v} |v\in V\} \in \mathbb{R}^{|V| \times d}$}
%}
%\sout{
%$\mathbf{X}^{V} = [X_{v_1}^\top, X_{v_2}^\top,...] \in \mathbb{R}^{|V| \times d}$
%($\mathbf{X}_v\in \mathbb{R}^d$ for the node $v$)
%is the node attribute matrix with feature dimension $d$, 
%and 
%$\mathbf{X}^{E}\in \mathbb{R}^{|E| \times d}$
%($\mathbf{X}_{uv} \in \mathbb{R}^d$ for edge $(u, v)$)
%is the edge attribute matrix.}
Each node $v$ and edge $(u, v)$ may be further equipped with attributes $\mathbf{X}_v \in \mathbb{R}^{d}$ and $\mathbf{X}_{uv} \in \mathbb{R}^{d}$.

In general, the learning on graph data first requires a graph encoder 
%$f_{\bm{\theta}}(\cdot)$ parameterized by $\bm{\theta}$ 
%{\color{blue}
$f_{\bm{\psi}, \bm{\theta}}(\cdot)$ with architectures $\bm{\psi}$ and parameters $\bm{\theta}$
%}
to map the original graph $G$ into the $d$-dimensional vector space: 
%\footnote{\# zhili: the following statement is not that accurate before, I've refined}
%{\color{brown}
%$\mathbf{H} = f_{\bm{\theta}}(G)$,
$f_{\bm{\psi}, \bm{\theta}}: G \rightarrow \{\mathbf{H} \}$,
where $\mathbf{H} \in \mathbb{R}^{d}$ can be the learned representation of single node/edge or the entire graph, depending on the prediction level of downstream tasks. 
%}
Then, $\mathbf{H}$ can be fed into an additional prediction head
%(e.g., linear classifier, multi-layer perceptron) \\
(e.g., linear classifier)
%$g_{\bm{\omega}}(\cdot)$ parameterized by $\bm{\omega}$ 
to 
predict the true label.
%{\color{blue}
Note that in the paper we discard the notation of prediction head for brievity.
%}
%$y$
%$y_{pred}=g_{\bm{\omega}}(\mathbf{H})$.
After that, the entire model 
%$g_{\bm{\omega}}(f_{\bm{\theta}}(\cdot))$ 
can be trained in an end-to-end manner 
supervised by task-specific labels via the labeld training dataset $\mathcal{D}^{(tra)}$:
%\begin{equation}
%\bm{\theta}^{\ast}, \bm{\omega}^{\ast} = \arg\min_{\bm{\theta}, \bm{\omega}} \mathcal{L}_{sup} (g_{\bm{\omega}}(f_{\bm{\theta}}(\cdot)); \mathcal{D}^{tra}),
%\label{eq: sup_training}
%\end{equation}
%{\color{blue}
\begin{equation}
	\bm{\theta}^{\ast} = \arg\min_{\bm{\theta}} \mathcal{L}_{sup} (f_{\bm{\psi}, \bm{\theta}}(\cdot); \mathcal{D}^{(tra)}),
	\label{eq: sup_training}
\end{equation}
where $\psi$ can be GCN,GIN and other architectures as introduced in Sec. \ref{sssec: related_gnn_2} and Sec. \ref{sssec: related_gnn_3}.
%}
$\mathcal{L}_{sup}(\cdot)$
is the supervised loss function (e.g., cross entropy).
The construction of 
$\mathcal{D}^{(tra)}$
%represents the graph data with labels
%and 
also depends on the downstream task, e.g., $\mathcal{D}^{(tra)}=\{(G,y)\}$ for the graph classification.
%For example, as an important and widely studied topic across physics, chemistry, and materials science domains, molecular property prediction (MPP) \cite{kearnes2016molecular} belongs to graph-level task. 
%Specifically, a molecule can be modeled as a graph, where nodes denote atoms, edges denote chemical bonds, and label can be related to toxicity or enzyme binding.
%Given a series of molecular graphs $\mathcal{G}= \{G_1, \dots, G_N\}$, only a subset of graphs $\mathcal{G}_{L} \in \mathcal{G}$ with corresponding properties $\mathcal{P}_{L}$ are known. The aim is to train the model $(f_{\bm{\theta}}, g_{\bm{\omega}})$ on labeled data $\mathcal{D}=(\mathcal{G}_{L}, \mathcal{P}_{L})$, so that it can be  employed to generate representations $\mathbf{H}_{G_i}$ for unlabeled molecular graphs $\mathcal{G}_{U}$ and infer their properties $\mathcal{P}_{U}$. 

\subsection{Graph Neural Networks (GNNs)}
\label{ssec: related_gnn}

%\footnote{maybe use $\bm{\phi}$ and $\bm{\Phi}$ to replace $\mathcal{M}$ ?\# zhili: we may refine notations for GNN model $f_{\bm{\theta}}(\cdot)$ in related work section, since we need to highlight its architecture $\mathcal{M}$ in method section. How about $f_{\bm{\mathcal{M}, \theta}}(\cdot)$? let us discuss}

%\subsubsection{General Message Passing}
%\label{sssec: related_gnn_1}

Recent years have witnessed the unprecedented success of GNNs for modeling graph data and dealing with various graph tasks. 
As one of powerful graph encoders $f_{\bm{\theta}}(\cdot)$, 
%GNNs typically
%leverage the graph topology $\mathbf{A}$ as well as features $\mathbf{X}^V$ and $\mathbf{X}^E$
%to achieve the graph representation $\mathbf{H}$.
GNNs
rely on graph topology and node/edge features 
to achieve the representation learning.
The majority of GNNs~\cite{kipf2016semi, hamilton2017inductive, velivckovic2017graph, xu2018powerful}
follow the message-passing paradigm \cite{gilmer2017neural} to 
learn representation $\mathbf{H}_v$ for given node $v$ by iteratively aggregating messages from its neighbors $N(v)$. 
Formally, the intra-layer message passing for $v$ can be formulated as:
%\begin{align}
%	&\mathbf{M}^{(k)}_{v} = AGG^{(k)}(
%	\{(\mathbf{H}^{(k-1)}_{u}, \mathbf{H}^{(k-1)}_{v}, \mathbf{X}_{uv}) | u \in N(v)\}),
%	\label{eq: gnn_agg}
%	\\
%	&\mathbf{H}^{(k)}_{v} = COMB^{(k)}(\mathbf{H}^{(k-1)}_{v}, \mathbf{M}^{(k)}_{v}),
%	\label{eq: gnn_comb}
%\end{align}
\begin{align}
	&\mathbf{M}_{v} \leftarrow AGG(
	\{(\mathbf{H}_{u}, \mathbf{H}_{v}, \mathbf{X}_{uv}) | u \in N(v)\}),
	\label{eq: gnn_agg}
	\\
	&\mathbf{H}_{v} \leftarrow COMB(\mathbf{H}_{v}, \mathbf{M}_{v}),
	\label{eq: gnn_comb}
\end{align}
where 
$AGG(\cdot)$ function aggregates neighboring messages to produce intermediate embedding $\mathbf{M}_{v}$,
%$\mathbf{M}_{v}$ is the intermediate representation collected from neighbors via aggregation function $AGG(\cdot)$, 
$COMB(\cdot)$ function combines information from neighbors and center node itself to update the representation $\mathbf{H}_{v}$.
The aggregation process iterates for $k$ times such that each node captures up to $k$-hop information in its learned representation $\mathbf{H}_{v}$.
Furthermore, for graph-level tasks,
%(e.g., molecular property prediction, where a molecular can be modeled as a graph $G$, and atoms and chemical bonds can be represented as nodes and edges), 
a permutation-invariant readout function $READOUT(\cdot)$ is further required to obtain graph-level representation $\mathbf{H}_{G}$ of the entire graph $G$:
\begin{equation}
	\mathbf{H}_{G} = READOUT(\{\mathbf{H}_{v}|v\in V\}).
	\label{eq: gnn_read}
\end{equation}
%where
%$READOUT(\cdot)$ can be simple non-parameterized function, e.g., sum pooling and mean pooling,
%or other more advanced methods \cite{cangea2018towards, ranjan2020asap, baek2021accurate}. 

%Obviously,
%after $k$ ($1\leq k \leq K$) iterations/layers of aggregation, 
%each node captures information from their $k$-hop neighbors. 
%Apart from intra-layer message passing. 
%For more flexible and expressive representation learning, recent GNN variants \cite{} further allow inter-layer message passing: 
%%$N(v)$, $AGG(\cdot)$, and $CONB(\cdot)$ are crucial functions in $f_{\theta}(\cdot)$ \cite{} to differentiate various GNNs and affect model expressiveness. 
%\begin{equation}
%	\mathbf{H}_{v} = FUSE(\mathbf{H}^{0}_{v}, \dots, \mathbf{H}^{K}_{v}),
%	\label{eq: gnn_fuse}
%\end{equation}
%where the fusion function $FUSE(\cdot)$ captures multi-scale informations from different intermediate layers to induce the final node representation $\mathbf{H}_{v}$. 
%In Eq. \eqref{eq: gnn_agg} and \eqref{eq: gnn_comb}, 

\subsubsection{Manual GNNs}
\label{sssec: related_gnn_2}
The majority of GNNs are specific instantiations of Eq. \eqref{eq: gnn_agg}, \eqref{eq: gnn_comb}, and \eqref{eq: gnn_read} and are designed manually. They mainly differ in several key functions, e.g., $N(v)$, $AGG(\cdot)$, $COMB(\cdot)$ and $READOUT(\cdot)$.
We next present several classic GNNs 
which are adopted in later experiments.
\begin{itemize}[leftmargin=*]
\item 
Graph Convolutional Network (GCN) \cite{kipf2016semi} 
adopts mean aggregation function $MEAN(\cdot)$
and non-linear activation $\sigma(\cdot)$, e.g., $ReLU(\cdot)$.
It proposes to transform intermediate embedding into representation
by trainable matrix
$\mathbf{W}$:
\[
\mathbf{M}_{v}\!\leftarrow\! MEAN(
\{\mathbf{H}_{u} | u \in  N(v) \cup \{v\}), \ \mathbf{H}_{v}\!\leftarrow\! \sigma(\mathbf{W}\mathbf{M}_{v}).
\]
%{\color{teal}
%\begin{align}
%&\mathbf{M}_{v} \leftarrow MEAN(
%\{\mathbf{H}_{u} | u \in  N(v) \cup \{v\}), 
%\label{eq: gcn_agg}
%\\
%&\mathbf{H}_{v} \leftarrow \sigma(\mathbf{W}\mathbf{M}_{v}).
%\label{eq: gcn_comb}
%\end{align}
%}	

\item 
GraphSAGE (SAGE) \cite{hamilton2017inductive} concatenates the intermediate embedding with the representation from previous iteration
before the transformation:
\[
\mathbf{M}_{v}\!\leftarrow\!MEAN(
\{\mathbf{H}_{u} | u \in N(v)\}),\ \mathbf{H}_{v}\!\leftarrow\!\sigma(\mathbf{W} [\mathbf{H}_{v}||\mathbf{M}_{v}]).
\]
%{\color{teal}
%\begin{align}
%&\mathbf{M}_{v} \leftarrow MEAN(
%\{\mathbf{H}_{u} | u \in N(v)\}),
%\label{eq: sage_agg}
%\\
%&\mathbf{H}_{v} \leftarrow \sigma(\mathbf{W} [\mathbf{H}_{v}||\mathbf{M}_{v}]).
%\label{eq: sage_comb}
%\end{align}
%}

\item Graph Isomorphism Network (GIN) \cite{xu2018powerful}, as one of the most expressive GNN architectures, 
adopts sum aggregation function $SUM(\cdot)$ and 
multi-layer perceptron $MLP(\cdot)$ 
to transform the combined messages.
The scalar $\epsilon$ is to balance the weights of messages from center node and its neighbors:
\[
\mathbf{M}_{v}\!\leftarrow\!SUM(
\{\mathbf{H}_{u} | u \in N(v)\}),\ \mathbf{H}_{v}\!\leftarrow\!MLP((1 + \epsilon) \mathbf{H}_{v} + \mathbf{M}_{v}).
\]
%\begin{align}
%&\mathbf{M}_{v} \leftarrow SUM(
%\{\mathbf{H}_{u} | u \in N(v)\}),
%\label{eq: gin_agg}
%\\
%&\mathbf{H}_{v} \leftarrow MLP((1 + \epsilon) \mathbf{H}_{v} + \mathbf{M}_{v}).
%\label{eq: gin_comb}
%\end{align}

\item 
Graph Attention Network (GAT) \cite{velivckovic2017graph} 
introduces the attentive function $ATT(\cdot)$ \cite{vaswani2017attention}
as its aggregation function:
\[
\mathbf{M}_{v} \leftarrow ATT(
\{\mathbf{H}_{u} | u \in  N(v)), \ \mathbf{H}_{v} \leftarrow  \sigma(\mathbf{W}\mathbf{M}_{v}).
\]
%\begin{align}
%&\mathbf{M}_{v} \leftarrow ATT(
%\{\mathbf{H}_{u} | u \in  N(v)),
%\label{eq: gat_agg}
%\\
%&\mathbf{H}_{v} \leftarrow  \sigma(\mathbf{W}\mathbf{M}_{v}).
%\label{eq: gat_comb}
%\end{align}

\end{itemize}

For the graph-level $READOUT(\cdot)$ function, it can be simple non-parameterized function, e.g., sum pooling and mean pooling,
or other more advanced methods \cite{cangea2018towards, ranjan2020asap, baek2021accurate}. 

%Furthermore, for the graph-level task,
%%(e.g., molecular property prediction, where a molecular can be modeled as a graph $G$, and atoms and chemical bonds can be represented as nodes and edges), 
%a permutation-invariant readout function $READOUT(\cdot)$ is further required to obtain graph-level representation $\mathbf{H}_{G}$ of the entire graph $G$:
%\begin{equation}
%	\mathbf{H}_{G} = READOUT(\{\mathbf{H}_{v}|v\in V\}),
%	\label{eq: gnn_read}
%\end{equation}
%where
%$READOUT(\cdot)$ can be simple non-parameterized function, e.g., sum pooling and mean pooling,
%or other more advanced methods \cite{cangea2018towards, ranjan2020asap, baek2021accurate}. 

\begin{figure}[!t]
	\centering
	\includegraphics[width=8.75cm]
	{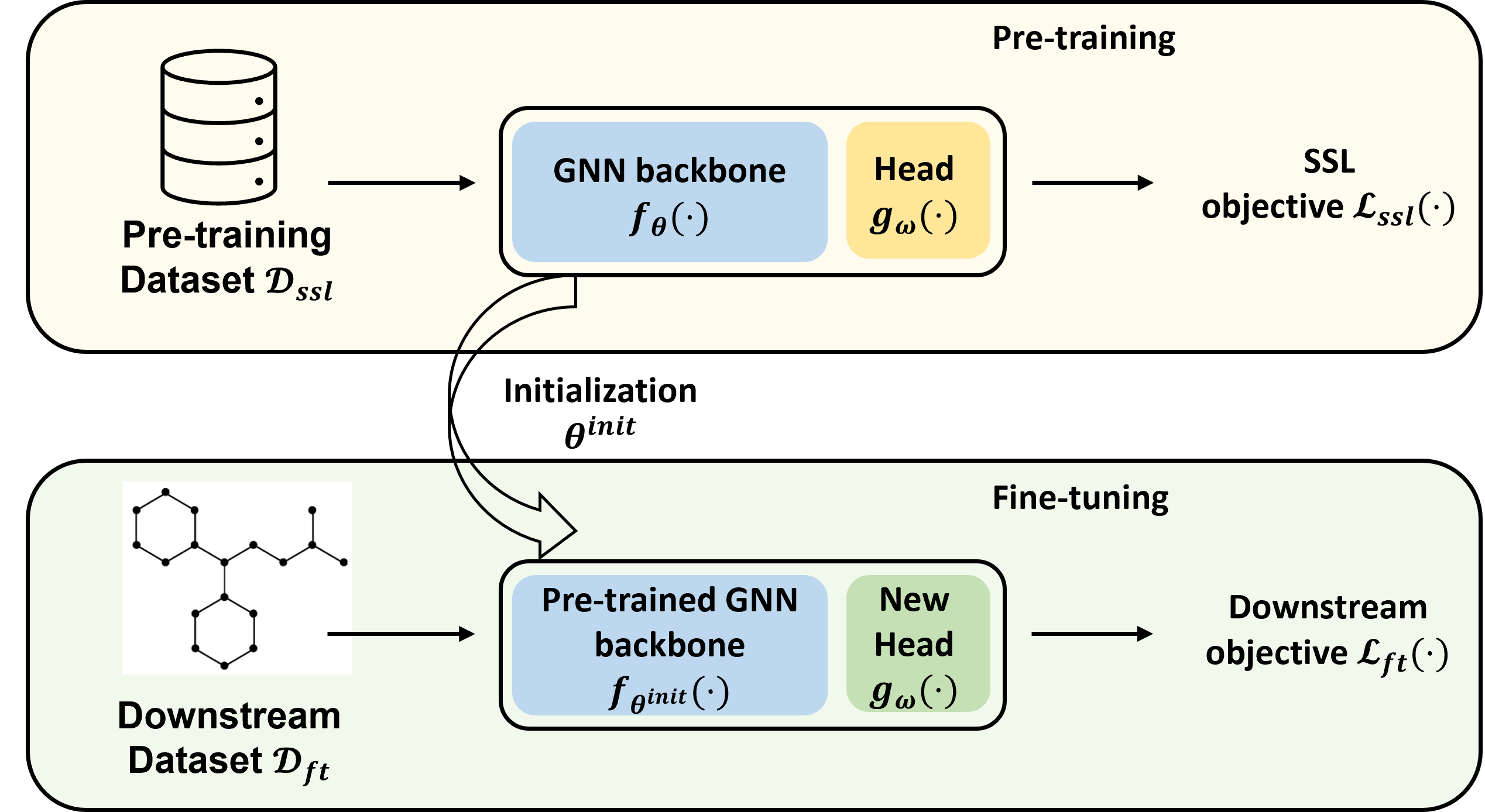}
	\caption{The illustration of overall GNN pre-training and fine-tuning framework.}
	\label{fig: gnn_pf}
\end{figure}

\subsubsection{Automated GNNs}
\label{sssec: related_gnn_3}

%{\color{blue}
To alleviate the extensive human labor in effective GNN architecture designs, recent efforts seek to automate this process and propose AutoGNNs \cite{gao2021graph, zhang2022pasca, wang2021autogel, oloulade2021graph, wang2023message, di2023message}. 
In general, AutoGNNs first design a unified GNN search space
$\mathcal{O}$
to cover key design functions in Eq. \eqref{eq: gnn_agg}, \eqref{eq: gnn_comb}, and \eqref{eq: gnn_read} (e.g., $AGG(\cdot)$, $COMB(\cdot)$) and promising candidates. They then adopt various search algorithms 
(e.g., differentiable methods \cite{liu2018darts}) 
to allow powerful GNNs to be discovered from the search space:
\begin{equation}
\bm{\psi}^{\ast}, \bm{\theta}^{\ast} = \arg\min_{\bm{\psi} \in \mathcal{O}, \bm{\theta}} \mathcal{L}_{sup} (f_{\bm{\psi}, \bm{\theta}}(\cdot); \mathcal{D}^{tra}).
\label{eq: autognn_training}
\end{equation}
Based on Eq.~\eqref{eq: autognn_training}, novel GNNs identified by AutoGNNs have demonstrated superior results than their manually-designed counterparts on many graph scenarios \cite{gao2021graph, zhang2022pasca, wang2021autogel, oloulade2021graph}.
%Unfortunately, existing AutoGNNs neglect important design dimensions for fine-tuning pre-trained GNNs, which makes them incapable to handle various downstream fine-tuning scenarios. 
%}

{\color{brown}
}

\begin{table*}[!htbp]
	\caption{Overview of common fine-tuning strategies in GNN and other domains.
		\textbf{Fine-tuning Scenarios} present the application scenario of the fine-tuning strategies,
		including 
		whether is designed for GNN or other neural networks
		and 
		which graph task the strategy can be applied to.
		\textbf{Fine-tuning Dimensions} presents what aspects they focus on during fine-tuning.
%		{\color{blue}
		\textbf{Automated} refers to whether they are capable of automatically designing the most suitable strategies to adaptively fine-tune the model for different downstream data. 
%		}
	}
	\label{tab: related_ft}
	\centering
	\begin{tabular}{cc|cc|cccc|c|c}
		\hline
		\hline
		\multicolumn{2}{l|}{\multirow{3}{*}{\bf Strategy Name}}
		&\multicolumn{2}{c|}{\bf Fine-tuning Scenarios}
		&\multicolumn{5}{c|}{\bf Fine-tuning Dimensions}
		&\multirow{3}{*}{\bf Automated}
		\\
		\cline{3-9}
		%		&&GNN &Other &Extra module &All tunable &Regularizer &
		&
		&\multirow{2}{*}{GNN/Other}
		&\multirow{2}{*}{Graph Task}
		&\multirow{2}{*}{\makecell{Model \\Architecture}} 
		&\multirow{2}{*}{Identity} 
		&\multirow{2}{*}{Fusion} 
		&\multirow{2}{*}{Readout} 
		&\multirow{2}{*}{\makecell{Model \\Weights}}&
		\\
		&&&&&&&&&\\
		\hline
		\multicolumn{2}{l|}{Vanilla Fine-Tuning (VFT)}&$\surd/\surd$ & Node/Edge/Graph  & $\times$ & $\times$ & $\times$ & $\times$ &$\surd$ &$\times$
		\\
		\hline
%		\multirow{7}{*}{\makecell{Regularized\\Fine-Tuning (RFT)}}
%		&\multicolumn{1}{|l|}{AUX-TS~\cite{han2021adaptive}} &$\surd/\times$&Node/Edge&$\times$ &$\times$ &$\times$ &$\times$ &\color{blue}$\surd$&\color{blue}$\times$
		\multirow{7}{*}{\makecell{Regularized\\Fine-Tuning (RFT)}}
		&\multicolumn{1}{|l|}{AUX-TS~\cite{han2021adaptive}} &$\surd/\times$&Node/Edge&$\times$ &$\times$ &$\times$ &$\times$ &$\surd$&$\times$
		\\
		\cline{2-10}
		&\multicolumn{1}{|l|}{WordReg~\cite{xia2022towards}} &$\surd/\times$ &Graph &$\times$ &$\times$ &$\times$ &$\times$ &$\surd$&$\times$
		\\
		\cline{2-10}
		&\multicolumn{1}{|l|}{GTOT-Tuning~\cite{zhang2022fine}} &$\surd/\times$ &Graph &$\times$ &$\times$ &$\times$ &$\times$ &$\surd$&$\times$
		\\
		\cline{2-10}
		&\multicolumn{1}{|l|}{$L^2$-SP~\cite{xuhong2018explicit}} & 
		$\times/\surd$ & - & $\times$ &$\times$& $\times$ & $\times$ &$\surd$ & $\times$
		\\
		\cline{2-10}
		&\multicolumn{1}{|l|}{DELTA~\cite{li2019delta}}& 
		$\times/\surd$ & - & $\times$&$\times$ &$\times$  &$\times$&$\surd$&$\times$
		\\
		\cline{2-10}
		&\multicolumn{1}{|l|}{BSS~\cite{chen2019catastrophic}} &
		$\times/\surd$ & - &$\times$ &$\times$&$\times$  &$\times$&$\surd$&$\times$
		\\
		\cline{2-10}
		&\multicolumn{1}{|l|}{StochNorm~\cite{kou2020stochastic}} &
		$\times/\surd$ & - &$\surd$ &$\times$&$\times$ &$\times$&$\surd$&$\times$
		\\
		\hline
		\multicolumn{2}{l|}{Feature Extractor (FE)}&$\times/\surd$ & - &$\times$&$\times$&$\times$ &$\times$&$\times$ &$\times$
		\\
		\hline
		\multicolumn{2}{l|}{Last-$k$ Tuning (LKT)}&$\times/\surd$  & - &$\times$&$\times$&$\times$&$\times$&$\surd$&$\times$
		\\
		\hline
		\multicolumn{2}{l|}{Adapter-Tuning (AT)}&$\times/\surd$ & - &$\surd$  &$\times$ &$\times$ &$\times$ &$\surd$&$\times$
		\\
		\hline
		\multicolumn{2}{l|}{S2PGNN}&$\surd/\times$ & Graph &$\surd$ &$\surd$ &$\surd$ &$\surd$&$\surd$&$\surd$
		\\
		\hline \hline
	\end{tabular}
\end{table*}

\subsection{Fine-tuning Pre-trained GNNs}
\label{ssec: related_pf}

%\sout{
%Despite the great success of GNNs, they are primarily trained via the supervised manner, which requires a large amount of task-specific labels to achieve satisfactory performances. However, in many practical graph scenarios (e.g., molecular property prediction in scientific domain), obtaining high-quality labels can be expensive and time-costly. Such data-scarcity issue greatly impedes the capacity of GNNs on graph data.}
%\footnote{
%	\# shimin: can be removed, we have clearly introduce this part in the introduction
%}

To relieve the reliance of GNNs on task-specific labels and improve their performances on scarcely-labeled graphs, 
%{\color{blue}
recent works \cite{hu2019strategies, zhang2021motif, you2020graph, xia2022pre}
%}
generalize the idea of self-supervised learning (SSL) \cite{liu2022graph} to graph data and propose to pre-train GNNs on 
%{\color{brown}
%recent efforts \cite{hu2019strategies, zhang2021motif, you2020graph} 
%try to generalize the self-supervised pre-training and fine-tuning (P\&F) paradigm to GNNs.
%}
%Generally,
%they first follow the self-supervised learning (SSL) way to 
%train a pre-trained GNN model on the 
large scale of unlabeled graph data 
$\mathcal{D}_{ssl}$:
\begin{align}
%\color{blue}
\bm{\theta}^{init} = \arg\min_{\bm{\theta}} \mathcal{L}_{ssl} (f_{\bm{\psi}, \bm{\theta}}(\cdot); \mathcal{D}_{ssl}),
\label{eq: related_pf_1}
\end{align}
where $\mathcal{L}_{ssl}(\cdot)$ is the loss function based on the adopted SSL pretext task.
Depends on the instantiation of $\mathcal{L}_{ssl}(\cdot)$, existing pre-trained GNNs can be roughly categorized into: 
%{\color{blue}
AutoEncoding~\cite{hamilton2017inductive, hou2022graphmae}, 
Autoregressive Modeling \cite{zhang2021motif}, 
Maksed Component Modeling \cite{hu2019strategies, xia2022mole},
Context Prediction \cite{hu2019strategies}, and 
Contrastive Learning \cite{velickovic2019deep, you2020graph, xu2021self, xia2022simgrace}.
More technical details can be found in Sec.~\ref{ssec: exp_b}.
%}

Then, 
to allow the knowledge transfer from large-scale $\mathcal{D}_{ssl}$ to small-scale downstream $\mathcal{D}_{ft}$,
they initialize the downstream GNN model 
%$f_{\bm{\psi}, \bm{\theta}}(\cdot)$
with 
backbone architectures and
pre-trained parameters $(\bm{\psi}, \bm{\theta}^{init})$.
%and perform fine-tuning with task-specific labels in $\mathcal{D}_{ft}$:
Then they perform model fine-tuning with task-specific labels in $\mathcal{D}_{ft}$ via the fine-tuning objective $\mathcal{L}_{ft}(\cdot)$:
%the down-streaming model is initialized by the pre-trained parameter $\bm{\theta}^{init}$ and fine-tuned with a small fine-tuning data $\mathcal{D}_{ft}$:
%{\color{red}
%%	\begin{align}
%%		\bm{\theta}^{\ast}, \bm{\omega}^{\ast} = \arg\min_{\bm{\theta}, \bm{\omega}} \mathcal{L}_{ft} [g_{\bm{\omega}}(f_{\bm{\theta}}(\cdot)); \mathcal{D}_{ft}],
%%		\label{eq: related_pf_2}
%%	\end{align}
%	\[
%	FT( g(f_{\theta, \omega}()); \textbf{parameter})
%	\]
%}
\begin{align}
%\color{blue}
	\bm{\theta}^{\ast} = \arg\min_{\bm{\theta}} \bm{\Phi}_{ft} [\mathcal{L}_{ft}(f_{\bm{\psi}, \bm{\theta}}(\cdot); \mathcal{D}_{ft})],
	\label{eq: related_pf_2}
\end{align}
%{\color{blue}
where the optimization for initialization model $f_{\bm{\psi}, \bm{\theta}^{init}}(\cdot)$ is guided by the fine-tuning strategy $\bm{\Phi}_{ft}$.
%where the initialization model $f_{\bm{\psi}, \bm{\theta}^{init}}(\cdot)$
%is optimized by
%the fine-tuning objective $\mathcal{L}_{ft}(\cdot)$, 
%and the entire process in Eq. \eqref{eq: related_pf_2} is 
%guided with the fine-tuning strategy $\bm{\Phi}_{ft}$.
In the next, 
we elaborate on existing methods of fine-tuning techniques $\bm{\Phi}_{ft}$ within and outside pre-trained GNNs. 
We further summarize the comparisons between the proposed S2PGNN with existing ones in Tab.~\ref{tab: related_ft}.

\begin{itemize}[leftmargin=*]
	\item 
	Vanilla Fine-Tuning (VFT): 
	VFT is probably the most prevalent tuning strategy to adopt pre-trained GNNs among existing literatures \cite{hu2019strategies, xia2022simgrace, zhang2021motif, you2020graph}. 
%	among existing GNN P\&F literatures \cite{hu2019strategies, xia2022simgrace, zhang2021motif, you2020graph}. 
	Under this schema, all parameters of the pre-trained GNN together with the new prediction head
%	$(\bm{\theta}_{init}, \bm{\omega})$
%	$(\bm{\theta}_{init}, \bm{\omega})$ of the GNNs
	are fine-tuned with the task-specific supervised loss $\mathcal{L}_{sup}(\cdot)$ (e.g., cross-entropy loss for classification task) for given downstream scenarios:
	\begin{equation}
%	\bm{\theta}^{\ast}, \bm{\omega}^{\ast} = \arg\min_{\bm{\theta}, \bm{\omega}} \mathcal{L}_{sup} ((g_{\bm{\omega}}(f_{\bm{\theta}}(\cdot)); \mathcal{D}_{ft}),
		\mathcal{L}_{ft}(\cdot) \equiv 	\mathcal{L}_{sup}(\cdot)
		\label{eq: st}
	\end{equation}
	
	\item 
	Regularized Fine-Tuning (RFT): 
	To prevent the overfitting and improve generalization, 
	RFT methods \cite{han2021adaptive, xia2022towards, zhang2022fine} 
	present improved fine-tuning strategies $\bm{\Phi}_{ft}$ by regularization. 
	They conduct constrained downstream adaptation
	to enforce the similarity of model parameters or learned representations.
%	before and after fine-tuning. 
%	to preserve the pre-trained model's learned parameters or features. 
	This is generally achieved by incorporating a regularization item $\mathcal{L}_{reg}(\cdot)$ with the original $\mathcal{L}_{sup}(\cdot)$:	
	\begin{equation}
		\mathcal{L}_{ft}(\cdot) =  \mathcal{L}_{sup}(\cdot)+ \mathcal{L}_{reg}(\cdot),
		\label{eq: ft_rt}
	\end{equation}
	where different RFT methods differ in the way to instantiate $\mathcal{L}_{reg}(\cdot)$.
	Among literatures,
%	{\color{brown}
%	+++move the introduction for AUX-TS \cite{han2021adaptive} and WordReg \cite{xia2022towards} from intro to here+++
	AUX-TS \cite{han2021adaptive} augments fine-tuning objectives with SSL objectives 
%	in an adaptive manner 
	so as to reduce the gap between two stages and improve results. 
%	improve the flexibility and effectiveness.
	WordReg \cite{xia2022towards} develops smoothness-inducing regularizer built on dropout \cite{srivastava2014dropout} to 
%	enforce the learned representations to be close 
	constrain representation distance induced by pre-trained and fine-tuned models. 
%	}
	GTOT-Tuning~\cite{zhang2022fine}
%	 is specifically designed to fine-tune pre-trained GNNs, which proposes to 
	uses optimal transport to align graph topologies of pre-trained and fine-tuned models while preserving learned representations.
%	(i.e., feature regularizer).
%	which considers the topology information in graph and presents an optimal transport-based feature regularizer. 
%	The authors argue that fine-tuning GNNs is challenging due to the complex and non-linear nature of the graph data, which can lead to overfitting and poor generalization. To address this issue, 
%	they propose a method that uses optimal transport to align the graph topologies of the pre-trained and fine-tuned models, while preserving the learned node representations.
\end{itemize}

\begin{figure*}[!t]
	\centering
	\subfloat[An example of vanilla fine-tuning strategy.]
	{\label{fig: a}
		\includegraphics[width=0.9\linewidth]
		{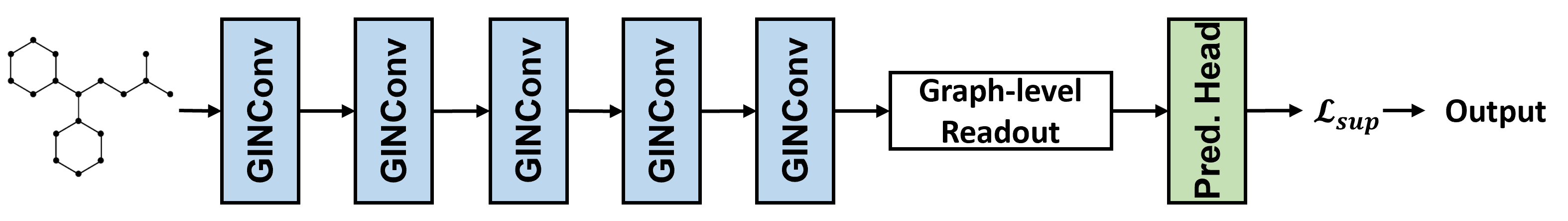}
	}\\
	\subfloat[An example of fine-tuning strategy with regularization.]
	{\label{fig: b}
		\includegraphics[width=0.9\linewidth]
		{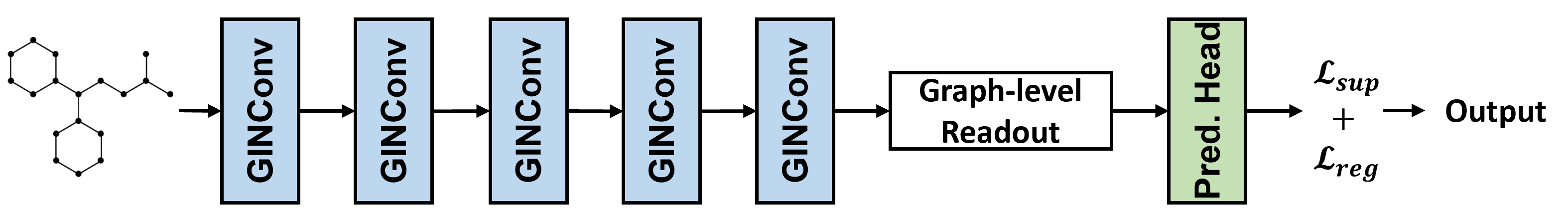}}\\	
	\caption{Illustration of GNN fine-tuning strategies (refer to Fig. \ref{fig: ours} for the legend).}
\end{figure*}

\newtheorem{remark}{Remark}
%{\color{blue}
\begin{remark}[Prompt Tuning]
%	PT approaches 
	Prompt Tuning in GNNs 
	\cite{sun2022gppt, liu2023graphprompt, sun2023all} propose to unify the pre-training and downstream graph tasks with the shared task template and leverage the prompt technique \cite{brown2020language} to prompt pre-trained knowledge for downstream learning.
	However, 
%	they mainly investigate how to unify tasks in two stages and  
%	the design of shared task template to unify pre-training and downstream tasks, 
	their methods may rely on the high relevance between the pre-training domain with the downstream one, which however may not hold in practical scenarios, especially when out-of-distribution predictions are demanded (e.g., molecular property prediction \cite{hu2019strategies}).
	Besides, they mainly investigate how to unify tasks in two stages and task template designs,   
	which is significantly different with our fine-tuning search problem (see Definition \ref{def: problem}). 
	Therefore, 
%	{\color{red}
%	However, 
%	they rely heavily on the high relevance between the pre-trained domain with the downstream one, which however may not hold. 
%	Besides, 
%	they focus on the design of shared task templates, which may also be limited.}
	due to the significantly different research scope with this work, prompting methods are excluded for comparison in Tab. \ref{tab: related_ft}.
%	which is a significantly different problem with ours.
%	Thus, due to the different scopes, PT methods are excluded for comparisons. 
	\label{remark: prompt}
\end{remark}

\begin{remark}[Fine-tuning Technique Outside GNNs]
Additionally, 
we further discuss and test several RFT methods (including 
$L^2$-SP~\cite{xuhong2018explicit},
DELTA~\cite{li2019delta},
BSS~\cite{chen2019catastrophic},
and StochNorm \cite{kou2020stochastic}) 
that are initially designed to fine-tune other types of deep models (e.g.,  CNNs) in the empirical study Sec.~\ref{ssec: exp_c}.
Besides, for more comprehensive exploration, we further cover other classic fine-tuning techniques that are originally designed outside GNN area, including:
Feature Extractor (FE) \cite{sharif2014cnn} that disables the fine-tuning to rely on parameter reuse, 
Last-$k$ Tuning (LKT) \cite{long2015learning} that freezes initial layers to fine-tune only last $k$ layers, and 
%Parameter-Efficient Tuning (PET) 
Adapter-Tuning (AT)
\cite{houlsby2019parameter} that is representative of parameter-efficient method which fine-tunes only a small number of extra parameters in Adapter modules.
%(e.g., Adapter \cite{houlsby2019parameter})
%Note that although existing GNN fine-tuning methods including RFT and PT try to improve the rudimentary SFT and present improved designs, there are still several obvious limitations which may hinder their downstream performances. 
\label{remark: ouside_gnn}
\end{remark}

\section{Methodology}
\label{sec: method}

%As introduced in Sec. \ref{sec: intro} and \ref{sec: related}, general GNNs follow the message passing framework to achieve representation learning. 
%When fine-tuning pre-trained GNNs to deal with various downstream scenarios, 
%proper adaption in GNN message functions (a.k.a., architectures) may be greatly demanded, so that the gap between pre-training and downstream stage can be bridged, and moreover, data/task-specific characteristics and requirements are better captured.
%Unfortunately, existing GNN fine-tuning works largely ignore the importance of GNN architecture and message function adaption, most of which simply use the fixed fine-tuning strategies for various downstream cases.  
%Obviously, existing solutions may be inflexible and insufficient.
%Therefore, in this work, we propose a novel GNN fine-tuning solution S2PGNN for improvements.

As introduced in Sec. \ref{sec: intro}, recently developed 
%pre-trained GNNs 
GNN pre-training techniques \cite{hu2019strategies, you2020graph, xu2021self, zhang2021motif, liu2021pre, xia2022simgrace, hou2022graphmae, xia2022pre}
have demonstrated promising to tackle the label-scarcity issue and improve the downstream learning.
Despite the emergence of various pre-training strategies,
how to fine-tune pre-trained GNNs for downstream scenarios, while important, is largely ignored in current literature.
%GNN pre-training works 
% mainly focus on the 
%largely ignore the design of powerful fine-tuning strategies to better transfer the pre-trained knowledge.
Only a few works start to investigate this direction,
but existing methods
% still suffer from several issues.
either have strong assumptions or overlook the data-aware issue behind various downstream domains.
%such as 
%practical infeasible,  
%limited effectiveness, and 
%not data-aware.
%as we illustrated in Sec. \ref{sec: intro}.

%{\color{blue}
To fully unleash the potential of pre-trained GNNs on various downstream datasets, 
we aim to design a fine-tuning searching strategy to automatically design suitable fine-tuning framework
for the 
given pre-trained GNN and downstream graph dataset.
However, it is a non-trivial task to achieve this search idea due to the undefined search space and large computational overhead for optimizing the search problem.
In this section, we first define a new problem of searching fine-tuning strategies for pre-trained GNNs.
We next propose an expressive search space, which enables powerful models to be searched.
Then, to reduce the search cost from the large and discrete space, we incorporate an efficient search algorithm and solve the search problem by differentiable optimization.
%}
%{\color{red}
%\sout{
%Specifically,
%we propose to automatically search the most suitable fine-tuning strategies given any pre-trained GNNs and downstream graph data. 
%To achieve this, 
%we first investigate existing fine-tuning literatures and provide a powerful fine-tuning search space that is suitable for pre-trained GNNs.
%However, the search space inevitably brings extra complexity and requires more computational overhead. 
%To address this, 
%we further incorporate an efficient search algorithm to achieve the searching goal.
%}}
%\footnote{\# shimin: discard this part for saving space}
%\sout{
%This section is organized as follows.
%In Sec. \ref{ssec: method_problem}, we formulate our search problem and provide the overall objective.
%In Sec. \ref{ssec: method_space}, we elabrate on the proposed GNN fine-tuning search space. 
%In Sec. \ref{ssec: method_alg}, we first conduct complexity analysis, and then present the efficient search algorithm and optimization. 
%}

%{\color{red}
%+++
%from this comment, we know the reviewer totally do not understand our contribution of our paper.
%Therefore, instead of just clarifying the Table 3, we need to carefully think how to formulate our contributions, and rewrite the space and algorithm
%+++
%}

\subsection{Problem Formulation}
\label{ssec: method_problem}

\newtheorem{Definition}{Definition}
\begin{Definition}[The Fine-tuning Strategy Search Problem for Pre-trained GNNs]
Given a pre-trained GNN model $f_{\bm{\psi}, \bm{\theta}}(\cdot)$ with 
architectures $\bm{\psi}$ and
parameters $\bm{}\theta$, 
%(i.e., $\bm{\Phi}_{ft} \sim \pi_{\bm{\alpha}}(\cdot)$)
and downstream graph dataset with split 
%$\mathcal{D}_{ft}=(\mathcal{D}_{ft}^{(tra)}, \mathcal{D}_{ft}^{(val)}, \mathcal{D}_{ft}^{(tst)})$.
$\mathcal{D}_{ft}=(\mathcal{D}_{ft}^{(tra)}, \mathcal{D}_{ft}^{(val)})$,
the automatic and data-aware GNN fine-tuning search problem in S2PGNN can be formally defined as
a bi-level optimization problem:
\begin{align}
\color{blue}
%	&
%	\color{red}
%	\bm{\alpha^{\ast}} = \arg\min_{\bm{\alpha}, \bm{\theta^{\ast}}} \mathbb{E}_{\bm{\Phi}_{ft} \sim \pi_{\bm{\alpha}}(\cdot)} \bm{\Phi}_{ft} [\mathcal{L}_{ft} (f_{\bm{\psi}, \bm{\theta^{\ast}}}(\cdot); \mathcal{D}_{ft}^{(val)})],
%	\\
	&
%	\color{blue}
	\bm{\Phi}_{ft}^* = \arg\min_{\bm{\Phi}_{ft}, \bm{\theta^{\ast}}} 
	\bm{\Phi}_{ft} [\mathcal{L}_{ft} (f_{\bm{\psi}, \bm{\theta^{\ast}}}(\cdot); \mathcal{D}_{ft}^{(val)})],
	\label{eq: obj_1}
	\\
	&
	{\rm s.t.} \quad \bm{\theta^{\ast}} =  \arg\min_{\bm{\theta}} \mathcal{L}_{ft} (f_{\bm{\psi}, \bm{\theta}}(\cdot); \mathcal{D}_{ft}^{(tra)}),
	\label{eq: obj_2}
\end{align}
where 
$\mathcal{L}_{ft}(\cdot)$ is the fine-tuning loss function, and 
$\bm{\Phi}_{ft}$ is the fine-tuning strategy that is to be searched. 
\label{def: problem}
\end{Definition}

%{\color{blue}
By comparing Eq.~\eqref{eq: autognn_training} and \eqref{eq: related_pf_2}  with above Eq.~\eqref{eq: obj_1},
it is intuitive that existing AutoGNNs focus on searching the GNN architecture $\bm{\psi}$ and optimizing $\bm{\theta}$. And existing pre-trained GNNs present a fixed fine-tuning strategy $\bm{\Phi}_{ft}$ and optimize the parameter $\bm{\theta}$. 
Both of them cannot be extended to solve the problem of searching fine-tuning strategies in Eq.~\eqref{eq: obj_1}.
Furthermore, it is hard to solve Eq.~\eqref{eq: obj_1} due to the unclear search space of $\bm{\Phi}_{ft}$ and large computational overhead of search algorithm.
More specifically, 
fine-tuning pre-trained GNNs is still an under-investigated problem and there lack a systematic and unified design space in existing literature.
Besides, it is very time-consuming to enumerate all possible choices of $\bm{\Phi}_{ft}$ because we need to train the model parameter $\bm{\theta}$ to convergence for each $\bm{\Phi}_{ft}$.
%}

\begin{figure*}[!t]
	\centering
	\subfloat
	{
		\includegraphics[width=1\textwidth]
		{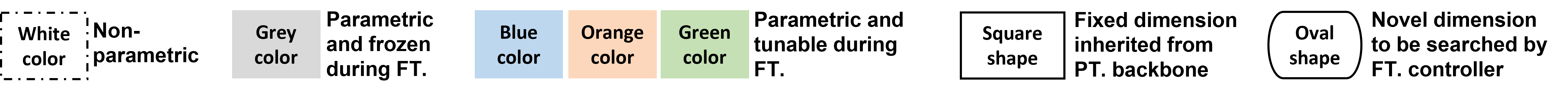}}\\	
	\vspace{+5px}
	\subfloat
	{
		\includegraphics[width=1\textwidth,height=0.125\textwidth]
		{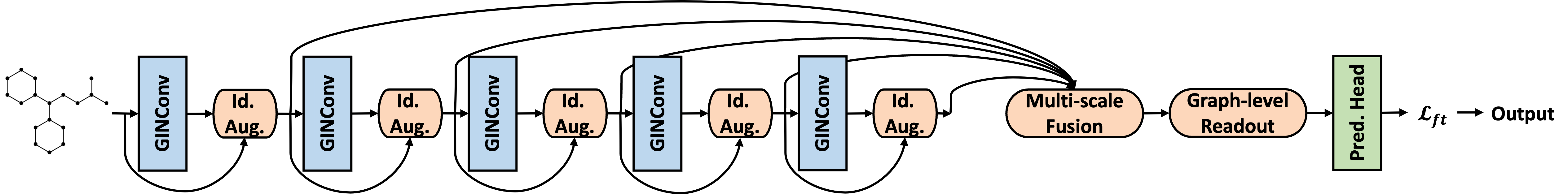}
	}\\
	\caption{Illustration to the framework  of S2PGNN built on top of pre-trained 5-layer GIN.
		The orange part indicates the search dimensions in S2PGNN.
		PT., FT., and Id.Aug. are abbreviations for pre-training, fine-tuning, and identity augmentation.}
	\label{fig: ours}
\end{figure*}

\subsection{GNN Fine-tuning Search Space}
\label{ssec: method_space}

To tackle the challenge of space design for GNN fine-tuning strategy, 
we provide a novel GNN fine-tuning search space from a brand new perspective, i.e., 
to incorporate the space of 
GNN structures with fine-tuning strategies. 
%which instantiates the fine-tuning search objective in Eq. \eqref{eq: obj_1} with a more concrete form:
%\begin{equation}
%	\bm{\phi}^* = \arg\min_{\bm{\phi}, \bm{\theta^{\ast}}} 
%	[\mathcal{L}_{ft} (f_{\bm{\phi}, \bm{\theta^{\ast}}}(\cdot); \mathcal{D}_{ft}^{(val)})],
%	\label{eq: obj_1_}
%\end{equation}}
More specifically, 
to ensure the improvement brought by searching fine-tuning strategies, 
we identify that the
%identity augmentation
%$\psi_{aug}(\cdot)$,
%multi-scale fusion
%$\psi_{fuse}(\cdot)$, and
%graph-level readout 
%$\psi_{read}(\cdot)$
backbone convolution
$\bm{\phi}_{conv}(\cdot)$,
identity augmentation
$\bm{\phi}_{id}(\cdot)$,
multi-scale fusion
$\bm{\phi}_{fuse}(\cdot)$, and
graph-level readout 
$\bm{\phi}_{read}(\cdot)$
for model structures 
%(which we will introduce in detail later)
are key functions to affect GNN fine-tuning results. 
Accordingly, 
the improved downstream message-passing with 
%incorporated 
key functions in S2PGNN 
%(see Fig. \ref{fig: ours})
%based on pre-trained GNNs
%As illustrated in Fig. \ref{fig: ours}, 
can be represented as:
\begin{equation}
	\left\{
	\begin{array}{rl}
		&\mathbf{Z}_{v} \leftarrow \bm{\phi}_{conv}(\{\mathbf{H}_{u}, \mathbf{H}_{v}, \mathbf{X}_{uv} | u \in N(v)\}),
		\label{eq: ours_mpnn_1}
		\\
		&\mathbf{H}_{v} \leftarrow \bm{\phi}_{id}(\mathbf{H}_{v}, \mathbf{Z}_{v}),  
		\quad 1\leq k \leq K,
	\end{array}
	\right.
\end{equation}
\vspace{-10px}
\begin{align}
	%	&\mathbf{H}_{v} = f_{fuse}(\{\mathbf{H}^{(1)}_{v}, \mathbf{H}^{(2)}_{v}, \dots, \mathbf{H}^{(K)}_{v}\}),
	&\mathbf{H}_{v} = \bm{\phi}_{fuse}(\{\mathbf{H}^{(k)}_{v} | 1\leq k \leq K\}),
	\label{eq: ours_mpnn_2}
	\\
	&\mathbf{H}_{G} = \bm{\phi}_{read}(\{\mathbf{H}_{v}|v\in V\}),
	\label{eq: ours_mpnn_3}
\end{align}
where 
%{\color{blue}
$ \bm{\phi}_{conv}(\cdot)$ summarizes the intra-layer message passing Eq. \eqref{eq: gnn_agg} and \eqref{eq: gnn_comb},
$\bm{\phi}_{id}(\cdot)$ augments the intra-layer message passing in pre-trained backbone $\bm{\phi}_{conv}(\cdot)$
with 
identity information from center node itself, 
$\bm{\phi}_{fuse}(\cdot)$ fuses the multi-scale information from different GNN layers $k$, 
$\bm{\phi}_{read}(\cdot)$ summarizes 
%information from every node 
node representations 
to yield the graph representation.
%}
%{\color{red}
%\footnote{
%	\# zhili: removed \# shimin: we should focus on explaining $\phi$,
%	it does not matter if you remove them
%}
%These incorporated (discrete) dimensions $\bm{\psi}$ are 
%associated with corresponding (continuous) parameters $\bm{\theta}$:
%\begin{equation}
%	\left\{
%	\begin{array}{rl}
%		&\bm{\psi} = \{\psi_{conv}(\cdot), \psi_{id}(\cdot), \psi_{fuse}(\cdot), \psi_{raad}(\cdot)\},
%		\label{eq: dimensions}
%		\\
%		&\bm{\theta} = \{\theta_{conv}, \theta_{id}, \theta_{fuse}, \theta_{raad}\}.
%	\end{array}
%	\right.
%\end{equation}
%}
%$\bm{\theta} = \{\theta_{conv}, \theta_{id}, \theta_{fuse}, \theta_{raad}\}$.
Based on Eq. \eqref{eq: ours_mpnn_1}, \eqref{eq: ours_mpnn_2}, and \eqref{eq: ours_mpnn_3}, 
the fine-tuning model structure is shown in Fig. \ref{fig: ours},
where we use the classic 5-layer GIN backbone \cite{hu2019strategies} as an example for demonstration.
%{\color{brown}
%The structure of fine-tuing model $f_{\bm{\psi}, \bm{\theta}}(\cdot)$ based on Eq. \eqref{eq: ours_mpnn_1}, \eqref{eq: ours_mpnn_2}, and \eqref{eq: ours_mpnn_3} are shown in Fig. \ref{fig: ours}, where we utilize the classic 5-layer GIN backbone \cite{hu2019strategies} as example for demonstration.}
Next, we 
elaborate on the  
%provide detailed 
illustrations of the proposed dimensions 
%$\psi_{id}(\cdot)$,
%$\psi_{fuse}(\cdot)$, and
%$\psi_{read}(\cdot)$
and their corresponding candidate-sets.

\subsubsection{Backbone Convolution $\bm{\phi}_{conv}(\cdot)$}
This dimension is the most basic building block in pre-trained GNNs.
We directly
%inherit 
transfer
its structure and parameters 
to the downstream model. 
%from pre-trained model.

\subsubsection{Identity Augmentation $\bm{\phi}_{id}(\cdot)$}
This dimension is important for data-aware fine-tuning due to several reasons. 
Firstly,
in some downstream data,
the identity information from node itself may be more essential than that aggregated from neighbors via $\bm{\phi}_{conv}(\cdot)$, especially when the neighboring messages can be missing, noisy, or unreliable. 
Secondly, 
the distinguishable information might be easily diluted and over-smoothed \cite{chen2020measuring} in some backbone choices     $\bm{\phi}_{conv}(\cdot)$, e.g., GCN. 
Thus, $\bm{\phi}_{id}(\cdot)$ may be beneficial to adjust the message flow in $\bm{\phi}_{conv}(\cdot)$.
%and preserve the distinguishable information for downstream data. 
We include candidates from:
{\color{brown}
%Augmenting the identity information $\mathbf{H}^{(k-1)}_{v}$ from center node itself can be indispensable for GNN fine-tuning due to several reasons.
%Firstly, in some downstream cases, node-specific information may be more important than messages from neighbors, since the latter sometimes can be missing, noisy, or unreliable in real-world datasets.
%Besides, in some GNN backbone architectures, node-specific information may be easy to lose and representations may be prone to become indistinguishable (a.k.a., over-smoothing \cite{chen2020measuring}) with the increase of GNN layers.
%This can happen if the adopted aggregation function in pre-trained backbones only considers the information of neighboring nodes (e.g., GCN).
%Unfortunately, existing GNN fine-tuning works tend to ignore the importance of preserving node-specific identity information and lack of corresponding designs 
%as shown in Tab. \ref{tab: related_ft}.
%Therefore,
%incorporating the identity augmentation during GNN fine-tuning may
%be powerful to prevent information loss and improve model robustness on downstream scenarios.
%Therefore, we propose to incorporate identity augmentation into the GNN fine-tuning design space and allows suitable augmentation to be adaptively searched from the following candidates in Eq. \eqref{eq: ours_mpnn_1}:
}

\begin{itemize}[leftmargin=*]
	\item \textbf{No augmentation.}
	We disable the identity augmentation and keep consistent as in pre-trained backbone with $zero\_aug$.
%	$\mathbf{Z}^{(k)}_{v} = \mathbf{H}^{(k)}_{v}$.
	
	\item \textbf{Additive augmentation.}
	We allow direct skip-connection \cite{li2019deepgcns}
	with $identity\_aug$ as
	$\mathbf{H}_{v} \leftarrow \mathbf{H}_{v} + \mathbf{Z}_{v}$.
	We also allow
	transformed augmentation with $trans\_aug$ as
	$\mathbf{H}_{v} \leftarrow g(\mathbf{H}_{v}) +  \mathbf{Z}_{v}$, where $g(\cdot)$ is a parameterized neural network with bottleneck architecture that maps $\mathbb{R}^{d} \rightarrow \mathbb{R}^{m} \rightarrow \mathbb{R}^{d}$, 
	where we enforce $m \ll d$ for parameter-efficient transformation similar to Adapter Tuning \cite{houlsby2019parameter}.
\end{itemize}

%\footnote{\color{brown}\# zhili: to modify this table: 1) add operator parameters in addition to existing categorical decisions, and 2) add the fixed intra-mpnn based on pre-trained backbones}
%\begin{table*}[!t]
%	\caption{The GNN fine-tuning design space in S2PGNN: design dimensions and candidate-sets.}
%	\label{tab: space}
%	\centering
%	%	\scalebox{1}{
%	\begin{tabular}{c|l|l|l}
%		\hline \hline
%		Type &\multicolumn{2}{c|}{Design Dimension $\bm{\phi}$} &Candidate-set $\mathcal{O}$ \\
%		\hline
%		\multirow{1}{*}{Intra-layer}
%		&Backbone Convolution
%		&$\bm{\phi}_{conv}(\cdot)$
%		&$\mathcal{O}_{conv}=\{pre\_trained\}$
%		\\
%		\hline 
%		\multirow{2}{*}{Inter-layer}
%		&Identity Augmentation
%		&$\bm{\phi}_{id}(\cdot)$
%		&$\mathcal{O}_{id}=\{zero\_aug, identity\_aug, trans\_aug\}$
%		\\ 
%		\cline{2-4}
%		&Multi-scale Fusion
%		&$\bm{\phi}_{fuse}(\cdot)$
%		&$\mathcal{O}_{fuse}=\{last, concat, max, mean, ppr, lstm, gpr\}$
%		\\
%		\hline 
%		\multirow{1}{*}{Graph-level}
%		&Graph-level Readout
%		&$\bm{\phi}_{read}(\cdot)$
%		&$\mathcal{O}_{read}=\{sum\_pooling, mean\_pooling, max\_pooling, set2set, sort\_pooling, neural\_pooling\}$
%		\\
%		\hline \hline
%	\end{tabular}
%	%	}
%	\vspace{-5px}
%\end{table*}

\begin{table*}[!t]
	\caption{The GNN fine-tuning design space in S2PGNN: design dimensions and candidate-sets.}
	\label{tab: space}
	\centering
	%	\scalebox{1}{
	\vspace{-5px}
		\begin{tabular}{l|l|l}
			\hline \hline
			\multicolumn{2}{l|}{Design Dimension } &\multicolumn{1}{l}{Candidate-set} \\
			\hline
			Backbone Convolution
			&$\bm{\phi}_{conv}(\cdot)$
			&$\mathcal{O}_{conv}=\{pre\_trained\}$
			\\
			\hline 
			Identity Augmentation
			&$\bm{\phi}_{id}(\cdot)$
			&$\mathcal{O}_{id}=\{zero\_aug, identity\_aug, trans\_aug\}$
			\\ 
			\hline 
			Multi-scale Fusion
			&$\bm{\phi}_{fuse}(\cdot)$
			&$\mathcal{O}_{fuse}=\{last, concat, max, mean, ppr, lstm, gpr\}$
			\\
			\hline 
			Graph-level Readout
			&$\bm{\phi}_{read}(\cdot)$
			&$\mathcal{O}_{read}=\{sum\_pooling, mean\_pooling, max\_pooling, set2set, sort\_pooling, neural\_pooling\}$
			\\
			\hline \hline
		\end{tabular}
		\vspace{-5px}
\end{table*}

\subsubsection{Multi-scale Fusion $\bm{\phi}_{fuse}(\cdot)$}
%The importance of this dimension for data-aware fine-tuning comes from several folds. 
Due to the diversity of downstream data structure (e.g., density, topology), 
the most suitable GNN layers required by different data may be highly data-specific \cite{liu2020towards, chien2020adaptive}. 
Besides, immediate representations in hidden layers of pre-trained GNNs often capture a spectrum of graph information with multiple scales, i.e., from local to global as layer increases \cite{xu2018representation}. 
Therefore, to make the fullest usage of pre-trained information,
we propose to fuse the multi-scale information from different layers
$\mathbf{H}_{v} = \sum_{k} w_{v}^{(k)} \mathbf{H}_{v}^{(k)}$
%to make the fullest usage of pre-trained information 
%and thereby
so as to allow the more adaptive and effective learning for downstream data.
We cover candidates from:

\begin{itemize}[leftmargin=*]
	\item \textbf{Non-parametric fusion.}
	They use simple non-parametric approaches to determine weights $w_{v}^{(k)}$ for fusion. 
	Among candidates listed in Tab. \ref{tab: space}, 
%	$last(\cdot)$, $concat(\cdot)$, $max(\cdot)$, $mean(\cdot)$, and $ppr(\cdot)$
	$last$, $concat$, $max$, $mean$, and $ppr$
	belong to this type.
%	where
	Specifically, 
	$last$ disables the fusion to directly take the single-scale representations from last layer,
%	the last-layer output as representations,
	$concat$ concatenates the multi-scale information for fusion,
%	conducts the concatenation of multi-scale information as
%	as
%	$\mathbf{H}_{v} = [\mathbf{Z}^{(1)}_{v}||\dots ||\mathbf{Z}^{(K)}_{v}]$,
	$max$ takes the maximum value from each channel to induce the fused representations,
	$mean$ assigns equal importance weights for information in each layer,
	and
	$ppr$ assigns decayed weights with Personalized PageRank \cite{gasteiger2018predict}.
	
	\item \textbf{Attentive fusion.}
	They allow the
%	Attentive fusion methods 
	adaptive importance weights $w_{v}^{(k)}$ via attention mechanisms,
%	which generally have 
	where $w_{v}^{(k)} \in [0, 1]$ and $\sum_{k} w_{v}^{(k)}=1$.
	We adopt the powerful $lstm$ fusion that is similar as in \cite{xu2018representation}.
%	to generate importance scores for the multi-scale information from multiple GNN layers, and then perform weighted summation to induce the final representations. 

	\item \textbf{Gated fusion.}
	Gated fusion methods use gating functions to selectively filter information at different layers. 
	We use $gpr$ method for this category similar as in \cite{chien2020adaptive}, which allows the adaptive scale as well as the sign of information, i.e., $w_{v}^{(k)} \in [-1, 1]$.
	
\end{itemize}

\subsubsection{Graph-level Readout $\bm{\phi}_{read}(\cdot)$}
This is the compulsory function for the downstream graph-level predictions. 
Different readout methods focus on the capture of information from different aspects, e.g., node features or graph topology \cite{grattarola2022understanding}.
%capturing different aspects of node features or graph topology \cite{grattarola2022understanding}.
Thus, 
downstream data with different structures and properties may have their data-specific requirements towards the effective readout. 
Candidates for this dimension include:
\begin{itemize}[leftmargin=*]
	\item \textbf{Simple readout.}
	They are parameter-free and computationally fast, which may be suitable for graph data where the overall graph topology is less important than individual node features.
	We include 
%	$sum\_pooling(\cdot)$, $mean\_pooling(\cdot)$, and $max\_pooling(\cdot)$
	$sum\_pooling$, $mean\_pooling$, and $max\_pooling$
	readouts for this type. 
	
	\item \textbf{Adaptive readout.}
	Adaptive methods aim to identify and capture most informative nodes or substructures into the graph representation via more sophisticated designs.
%	Other readout methods, such as adaptive readout, 
%	focus on identifying and capturing the most informative nodes or substructures via more sophisticated designs.
%	They may be more effective for scenarios where specific nodes or substructures in the graph are more crucial.
	We review existing literatures to cover powerful candidates 
	$set2set$ \cite{vinyals2015order}, $sort\_pooling$ \cite{zhang2018end}, $multiset\_pooling$ \cite{baek2021accurate}, and $neural\_pooling$ \cite{buterez2022graph}
	for this category.
%	{\color{brown}
%	Other than the simple methods, there are more sophisticated design choices available in literatures, such as adaptive methods.
%	This type of methods usually introduce trainable operator parameters so that they can learn to identify the most important/informative nodes or substructures in the graph. 
%	They may be more effective for scenarios where specific nodes or substructures in the graph are more crucial.
%	}
\end{itemize}

\begin{remark}[Space Complexity of $\bm{\Phi}_{ft}$]	
%{\color{blue}
As shown in Fig.~\ref{fig: ours} and summarized in Tab. \ref{tab: space}, the overall complexity of the proposed GNN fine-tuning strategy space $\bm{\Phi}_{ft}$ in S2PGNN equals to the Cartesian product of the size of all involved dimensions, 
i.e., $O(|\mathcal{O}_{conv}|^{K}\cdot |\mathcal{O}_{id}|^{K} \cdot |\mathcal{O}_{fuse}| \cdot |\mathcal{O}_{read}|)$.
%$= 1^5 \times 3^5 \times 7 \times 6 =10206$.
%}
For the illustration in Fig.~\ref{fig: ours}, the search space built on the 5-layer GIN will have 10,206 candidate fine-tuning strategies.
Because each candidate will require to train the GNN parameter to convergence,
we cannot simply use a brute-force algorithm to test all possible candidates and train thousands of GNNs, which can lead to enormous computational overhead.
%}
\label{remark: complexity}
\end{remark}

\subsection{GNN Fine-tuning Search Algorithm}
\label{ssec: method_alg}

%{\color{blue}
Recall that as introduced in Sec. \ref{ssec: method_problem}, 
the GNN fine-tuning strategy search problem is formally defined as Eq. \eqref{eq: obj_1} and \eqref{eq: obj_2}.
%where we aim to search the optimial GNN fine-tuning strategy $\bm{\phi}$ such that it can 
%minimize the expected fine-tuning loss (maximize the expected performance) derived from the model fine-tuned via this searched strategy.
To solve this problem, 
we need to compute the gradients of 
GNN fine-tuning strategy $\nabla_{\bm{\phi}}\mathcal{L}_{ft}$ and 
GNN model weights $\nabla_{\bm{\theta}}\mathcal{L}_{ft}$.
Computing $\nabla_{\bm{\theta}}\mathcal{L}_{ft}$ is simple since $\bm{\theta}$ is continuous.
Unfortunately,
the GNN fine-tuning strategy $\bm{\phi}$ is categorical choices (see Tab. \ref{tab: space}) and thus discrete, thereby $\nabla_{\bm{\phi}}\mathcal{L}_{ft}$ does not exist.
Thus, we propose to train a controller $p_{\bm{\alpha}}(\phi)$ that parameterized by $\bm{\alpha}$ and hope this controller can sample better fine-tuning strategy $\phi$ after achieving the performance of the sampled $\phi$.
Therefore, we reformulate the problem into:
\begin{align}
	&\bm{\alpha^{\ast}} = \arg\min_{\bm{\alpha}, \bm{\theta^{\ast}}} \mathbb{E}_{\bm{\phi} \sim p_{\bm{\alpha}}(\bm{\phi})} [\mathcal{L}_{ft} (f_{\bm{\phi}, \bm{\theta^{\ast}}}(\cdot); \mathcal{D}_{ft}^{(val)})],
	\label{eq: real_obj_1}\\
	&{\rm s.t.} \quad \bm{\theta^{\ast}} =  \arg\min_{\bm{\theta}} \mathcal{L}_{ft} (f_{\bm{\phi}, \bm{\theta}}(\cdot); \mathcal{D}_{ft}^{(tra)}),
	\label{eq: real_obj_2}
\end{align}
where $\bm{\alpha}$ is the continuous controller parameter that controls the selection of GNN fine-tuning strategy $\bm{\phi}$,
$\bm{\phi} \sim p_{\bm{\alpha}}(\bm{\phi})$ represents a GNN fine-tuning strategy $\bm{\phi}$ being sampled from the distribution $p_{\bm{\alpha}}(\bm{\phi})$ parameterized by $\bm{\alpha}$,
and
$\mathbb{E}[\cdot]$ is the expectation function.
We aim to optimize $\bm{\alpha}$ such that it can yield the optimal fine-tuning strategy $\bm{\phi}^{\ast} \sim p_{\bm{\alpha}^{\ast}}(\bm{\phi})$.
In the following, we demonstrate the concrete algorithm to solve Eq. \eqref{eq: real_obj_1} and \eqref{eq: real_obj_2}
in Sec. \ref{sssec: method_alg_update_strategy} and Sec. \ref{sssec: method_alg_update_weight}.

\subsubsection{Update Fine-tuning Strategy by Re-parameterization}
%\subsubsection{Update Fine-tuning Strategy by Continuous Relaxation}
\label{sssec: method_alg_update_strategy}

We propose to employ dimension-specific controllers 
$\bm{\alpha} = \{\bm{\alpha}_{conv}, \bm{\alpha}_{id}, \bm{\alpha}_{fuse}, \bm{\alpha}_{raad}\}$ 
to guide the GNN fine-tuning strategy search 
from $\mathcal{O}= \{\mathcal{O}_{conv}, \mathcal{O}_{id}, \mathcal{O}_{fuse}, \mathcal{O}_{raad}\}$ 
for respective dimensions (see Sec. \ref{ssec: method_space} and Tab. \ref{tab: space}),
%i.e., 
%to controll the selection from
%$\mathcal{O}= \{\mathcal{O}_{conv}, \mathcal{O}_{id}, \mathcal{O}_{fuse}, \mathcal{O}_{raad}\}$ 
%for dimensions
%$\bm{\phi} = \{\bm{\phi}_{conv}(\cdot), \bm{\phi}_{id}(\cdot), \bm{\phi}_{fuse}(\cdot), \bm{\phi}_{raad}(\cdot)\}$,
where $\bm{\alpha}_{dim} \in \mathbb{R}^{|\mathcal{O}_{dim}|}$ 
for $\forall dim \in \{conv, id, fuse, read\}$.
%Note that in the following, the notation of $dim$ is discarded for brevity.)
Then,
for the given fine-tuning dimension 
%$dim$ 
with candidate-set 
%$\mathcal{O}_{dim}$,
$\mathcal{O}$
(note that we discard the notation of dimension $dim$ for brevity), 
let the one-hot vector 
%$\bm{\phi} \in \mathbb{R}^{|\mathcal{O}|}$ 
$\bm{\phi} \sim p_{\bm{\alpha}}(\bm{\phi}) \in \mathbb{R}^{|\mathcal{O}|}$
represent the sampled strategy,
let $\mathbf{Z}_{in}$ be the intermediate representation that to be fed into this dimension, 
%$\mathbf{Z}_{in}$ as input for this fine-tuning dimension, 
then its output is calculated as
$\mathbf{Z}_{out} = \sum\nolimits_{i=1}^{|\mathcal{O}|} \bm{\phi}[i] \cdot \mathcal{O}[i](\mathbf{Z}_{in})$.
%$\bm{\phi}[i] \in \{0,1\}$ and $\sum\nolimits_{i=1}^{|\mathcal{O}|} \bm{\phi}[i]=1$.
However, the sampling process 
$\bm{\phi} \sim p_{\bm{\alpha}}(\bm{\phi})$ 
is discrete, 
which makes the derivative
$\nabla_{\bm{\alpha}}\mathbf{Z}_{out}$ undefined and 
the gradient 
%w.r.t. $\bm{\alpha}$
$\nabla_{\bm{\alpha}}\mathcal{L}_{ft}$
non-existent. 
%entire framework non-differentiable and therefore cannot 
%backpropagate gradients to achieve the optimization.

%{\color{blue}
To address this issue for
optimizing strategy controller parameters $\bm{\alpha}$ in Eq. \eqref{eq: real_obj_1},
we leverage the re-parameterization trick in \cite{jang2016categorical, maddison2016concrete} to 
relax the discrete strategy sampling 
%$\bm{\phi} \sim p_{\bm{\alpha}}(\bm{\phi})$
to be continuous and differentiable via $\bm{\phi}=g_{\bm{\alpha}}(\bm{U})$:
%conduct continuous relaxation towards the discrete $\bm{\phi} \sim \pi_{\bm{\alpha}}(\cdot)$ as:
%\begin{align}
%%	=\frac
%%	{\exp ((\log \bm{\alpha}[i] - \log(-\log(\bm{U}[i])))/ \tau)}
%%	{\sum\nolimits_{j=1}\exp ((\log \bm{\alpha}^{\mathcal{O}}[j] - \log(-\log(\bm{U}[j])))/ \tau)},
%	\bm{\phi}[i]
%	&= g_{\bm{\alpha}}(\bm{G}[i])\\
%	&=\frac
%	{\exp ((\log \bm{\alpha}[i] + \bm{G}[i])/ \tau)}
%	{\sum\nolimits_{j=1}^{|\mathcal{O}|}\exp ((\log \bm{\alpha}[j] + \bm{G}[j])/ \tau)},
%	\label{eq: softmax}
%\end{align}
\begin{align}
%	\bm{\phi}[i]
%	&= g_{\bm{\alpha}}(\bm{G}[i])\\
%	&=\frac
%	{\exp ((\log \bm{\alpha}[i] + \bm{G}[i])/ \tau)}
%	{\sum\nolimits_{j=1}^{|\mathcal{O}|}\exp ((\log \bm{\alpha}[j] + \bm{G}[j])/ \tau)},
%	\bm{\phi}[i]
%	\!\!=
	\!\! g_{\bm{\alpha}}(\bm{U}[i])
	\!\!=\!\! \frac
	{\exp ((\log \bm{\alpha}[i]-\log(-\log(\bm{U}[i])))/ \tau)}
	{\sum\nolimits_{j=1}^{|\mathcal{O}|}\!\exp ((\log \bm{\alpha}[j]\!-\!\log(-\log(\bm{U}[i])))/ \tau)},
	\label{eq: softmax}
\end{align}
where 
%$\bm{G}[i]= -\log(-\log(\bm{U}[i]))$,
%is the \textit{Gumbel} random variable \cite{jang2016categorical, maddison2016concrete},
%i,.e., 
%$\bm{G}[i]= -\log(-\log(\bm{U}[i]))$,
%%$\bm{U} \in \mathbb{R}^{|\mathcal{O}|}$ with 
%where
%and
$\bm{U}[i] \sim Uniform(0, 1)$ is sampled from the uniform distribution, and
$\tau$ is the temperature that controls the discreteness of softmax output.
In this way, 
the relaxed
%$\bm{\phi}[i]=g_{\bm{\alpha}}(\bm{U}[i])$
strategy $g_{\bm{\alpha}}(\bm{U})$
is differentiable w.r.t. $\bm{\alpha}$. 
Then, the gradient 
$\nabla_{\bm{\alpha}}\mathcal{L}_{ft}$
for solving Eq. \eqref{eq: real_obj_1}
can be computed as:
\begin{align}
&\nabla_{\bm{\alpha}} \mathbb{E}_{\bm{\phi} \sim p_{\bm{\alpha}}(\bm{\phi})} [\mathcal{L}_{ft} (\bm{\phi}, \bm{\theta^{\ast}}; \mathcal{D}_{ft}^{(val)})]\nonumber\\
= \ &\nabla_{\bm{\alpha}} \mathbb{E}_{\bm{U} \sim p(\bm{U})}[\mathcal{L}_{ft}(g_{\bm{\alpha}}(\bm{U}),\bm{\theta^{\ast}}; \mathcal{D}_{ft}^{(val)})]\nonumber\\
%=\ &\nabla_{\bm{\alpha}} \int p(\bm{U})\mathcal{L}_{ft}(g_{\bm{\alpha}}(\bm{U}),\bm{\theta^{\ast}}; \mathcal{D}_{ft}^{(val)})d\bm{U}\nonumber\\
%=\ &\int p(\bm{U})\nabla_{\bm{\alpha}}\mathcal{L}_{ft}(g_{\bm{\alpha}}(\bm{U}),\bm{\theta^{\ast}}; \mathcal{D}_{ft}^{(val)})d\bm{U}\nonumber\\
=\ &\mathbb{E}_{\bm{U} \sim p(\bm{U})} [\nabla_{\bm{\alpha}}\mathcal{L}_{ft}(g_{\bm{\alpha}}(\bm{U}),\bm{\theta^{\ast}}; \mathcal{D}_{ft}^{(val)})]\nonumber\\
=\ &\mathbb{E}_{\bm{U} \sim p(\bm{U})} [\mathcal{L}_{ft}^{'}(g_{\bm{\alpha}}(\bm{U}),\bm{\theta^{\ast}}; \mathcal{D}_{ft}^{(val)})\nabla_{\bm{\alpha}}g_{\bm{\alpha}}(\bm{U})],
\label{eq: gradient_alpha}
\end{align}	
where $\nabla_{\bm{\alpha}}g_{\bm{\alpha}}(\bm{U})$ can be computed because $g_{\bm{\alpha}}(\bm{U})$ presented in Eq. \eqref{eq: softmax} is differentiable
and Eq. \eqref{eq: gradient_alpha} can be easily approximated by Monte-Carlo (MC) sampling \cite{metropolis1949monte}.
Note that $\tau \rightarrow 0$ makes the continuous output in Eq. \eqref{eq: softmax} almost indistinguishable with the discrete one-hot vector,
which thereby ensures the relaxation in Eq. \eqref{eq: softmax} to be unbiased once converged.

\subsubsection{Update GNN Weights by Weight-sharing}
\label{sssec: method_alg_update_weight}

The gradient computation for high-level Eq. \eqref{eq: real_obj_1} requires frequent performance evaluation of sampled GNN fine-tuning strategy,
which leads to heavy computational cost. 
To alleviate this issue and accelerating search, 
we suggest the weight-sharing \cite{pham2018efficient} in low-level Eq. \eqref{eq: real_obj_2}. 
Specifically, 
we evaluate every sampled strategy $\bm{\phi}$ (Eq. \eqref{eq: softmax}) based on the shared GNN models weights $\bm{\theta}$, thereby we avoid repeatedly training the GNN model weights from scratch. 
The gradient w.r.t. shared model weights $\nabla_{\bm{\theta}}\mathcal{L}_{ft}$ for solving Eq. \eqref{eq: real_obj_2} can be calculated as:
%\footnote{\color{brown}\# zhili: to modify since i miss the derivative in this equarion}
\begin{align}
	\nabla_{\bm{\theta}} \mathcal{L}_{ft} (\bm{\phi},\!\bm{\theta};\! \mathcal{D}_{ft}^{(tra)})
	\!=\!
	\frac{1}{|\mathcal{D}_{ft}^{(tra)}|} \!\! \sum_{(x,y) \in \mathcal{D}_{ft}^{(tra)}} \!\!\! \nabla_{\bm{\theta}} \ell(\bm{\phi},\!\bm{\theta};\!(x,y)),
	\nonumber
	%\label{eq: gradient_theta}
\end{align}	
where $\ell(\cdot)$ is the loss for each labeled data instance $(x, y)$. 
Empirically, we leverage the cross-entropy loss for graph classification task and MSE loss for graph regression task.

\begin{table*}[!t]
	\caption{Summary of downstream GNN fine-tuning datasets $\mathcal{D}_{ft}$.}
	\label{tab: exp_data}
	\centering
	\setlength\tabcolsep{10pt}
	\vspace{-5px}
	\begin{tabular}{l|lllll}
		\hline \hline
		\bf Dataset & \bf \#Molecules & \bf \#Tasks & \bf Task Type & \bf Metric & \bf Domain
		\\ \hline \hline
		BBBP &2039 &1 &Classification &ROC-AUC (\%) ($\uparrow$) &Pharmacology	
		\\ \hline
		Tox21 &7831 &12 &Classification &ROC-AUC (\%) ($\uparrow$) &Pharmacology	
		\\ \hline
		ToxCast &8575 &617 &Classification &ROC-AUC (\%) ($\uparrow$) &Pharmacology	
		\\ \hline
		SIDER &1427 &27 &Classification &ROC-AUC (\%) ($\uparrow$) &Pharmacology	
		\\ \hline
		ClinTox &1478 &2 &Classification &ROC-AUC (\%) ($\uparrow$) &Pharmacology	
		\\ \hline
		BACE &1513 &1 &Classification &ROC-AUC (\%) ($\uparrow$) &Biophysics	
		\\ \hline
		ESOL &1128 &1 &Regression &RMSE ($\downarrow$) &Physical Chemistry	
		\\ \hline
		Lipophilicity (Lipo) &4200 &1 &Regression &RMSE ($\downarrow$) &Physical Chemistry	
		\\ \hline \hline
	\end{tabular}
	\vspace{-5px}
\end{table*}

\section{Experiments}
\label{sec: exp}

As presented in Sec.~\ref{sec: related},
the GNN pre-training approach
generally contains a GNN backbone model (e.g., GCN~\cite{kipf2016semi}, SAGE~\cite{hamilton2017inductive}),
a GNN pre-training strategy (e.g., MCM~\cite{hu2019strategies,xia2022mole}, CL~\cite{velickovic2019deep}) to transfer knowledge from a large scale of unlabeled graph data $\mathcal{D}_{ssl}$,
a GNN fine-tuning strategy (e.g., ST~\cite{hu2019strategies}, RT~\cite{zhang2022fine}) 
to adapt the pre-trained GNNs to the domain-specific data with labels $\mathcal{D}_{ft}$.
In this paper, we mainly investigate the improvement from the perspective of automatically designing a suitable fine-tuning strategy for the given data $\mathcal{D}_{ft}$.
Thus, to validate the effectiveness of the proposed S2PGNN,
we need to answer the following questions:
\begin{itemize}[leftmargin=*]
	\item \textbf{Q1}: 
	Can S2PGNN 
	be built on top of different GNN backbone and pre-training methods and consistently improve their performance?
	(see Sec. \ref{ssec: exp_b} and Sec. \ref{ssec: exp_f})
	
	\item \textbf{Q2}: 
	On the same configuration of GNN backbone model and pre-training strategy,
	can S2PGNN 
	be more effective than other GNN fine-tuning strategies?
	(see Sec. \ref{ssec: exp_c})
	
	\item 
	\textbf{Q3}:
	As discussed in Sec.~\ref{ssec: related_pf},
	there are several classic fine-tuning strategies in other domains (see Remark \ref{remark: ouside_gnn})
	that are not included in our search space.	
	Will these models perform well on the GNN area?
	(see Sec. \ref{ssec: exp_c})
	
	\item 
	\textbf{Q4}: 
	What are the effects of each design dimension in S2PGNN? 
	(see Sec. \ref{ssec: exp_e})
	
	\item 
	\textbf{Q5}: 
	Is there a risk that the fine-tuning search method will consume more computing resources to achieve improved performance?
	(see Sec. \ref{ssec: exp_g})

\end{itemize}

\subsection{Experimental Settings}
\label{ssec: exp_setting}

S2PGNN 
\footnote{Code and data are available at 
%\url{https://anonymous.4open.science/r/code_icde2024-A9CB}.
\url{https://github.com/zwangeo/icde2024}.}
is implemented based on PyTorch \cite{paszke2019pytorch} and PyTorch Geometric \cite{fey2019fast} libraries. All experiments are conducted with one single NVIDIA Tesla V100 GPU.
%\footnote{\# zhili: may refer to "Geometry-enhanced molecular representation learning for property prediction" and "MOLE-BERT: RETHINKING PRE-TRAINING GRAPH NEURAL NETWORKS FOR MOLECULES" to adjust exp setting part}

\subsubsection{GNN backbone models}

For GNN backbone architectures, 
we mainly adopt 
classic and promising GNNs in recent years,
including 
(5-layer) 
GCN \cite{kipf2016semi}, 
SAGE \cite{hamilton2017inductive}, 
Graph Isomorphism Network (GIN) \cite{xu2018powerful}, 
and GAT \cite{velivckovic2017graph} (see more details in Sec.~\ref{ssec: related_gnn}). 
But due to the limited space, 
the experiments
in sections
Sec.~\ref{ssec: exp_b}, Sec.~\ref{ssec: exp_c}, Sec.~\ref{ssec: exp_e}, and Sec.~\ref{ssec: exp_g}
are conducted on GIN.
And we report 
the performance on
other backbone models in  Sec. \ref{ssec: exp_f}.
%	For GNN backbone architectures, 	
%	we mainly explore on Graph Isomorphism Network (GIN) \cite{xu2018powerful}, due to its expressiveness on graph-level tasks and widespread usage among related works. To be more comprehensive, we also evaluate S2PGNN with other popular GNN architectures as backbones (in Sec. \ref{sssec: exp_b2}): GCN \cite{kipf2016semi}, SAGE \cite{hamilton2017inductive}, and GAT \cite{velivckovic2017graph}.

\subsubsection{GNN pre-training methods and datasets $\mathcal{D}_{ssl}$}
\label{sssec: exp_setting_base}

To demonstrate the effectiveness of S2PGNN,
%	and readily pluggable to any advanced ,
we implement S2PGNN on top of 10 well-known 
and publicly available
pre-training methods,
including
Infomax \cite{velickovic2019deep}, 
EdgePred \cite{hamilton2017inductive}, 
ContextPred \cite{hu2019strategies}, 
AttrMasking \cite{hu2019strategies},
GraphCL \cite{you2020graph}, 
GraphLoG \cite{xu2021self}, 
MGSSL \cite{zhang2021motif},
SimGRACE \cite{xia2022simgrace}, 
GraphMAE \cite{hou2022graphmae}, and 
Mole-BERT \cite{xia2022mole}.
As summarized in Tab. \ref{tab: exp_pretrain_obj},
they 
cover a wide spectrum of various 
SSL strategies $\mathcal{L}_{ssl}(\cdot)$
(see more details in Sec. \ref{ssec: exp_b}). 
Due to the space limit, the fine-tuning experiments in Sec. \ref{ssec: exp_b} are conducted on top of all 10 pre-trained models, and experiments in Sec.~\ref{ssec: exp_c}, Sec.~\ref{ssec: exp_e}, Sec.~\ref{ssec: exp_g}, and Sec.~\ref{ssec: exp_g} are built on top of the pioneering pre-training work ContextPred \cite{hu2019strategies}.

In this paper, 
we follow the literature
to adopt ZINC15 (250K) for MGSSL~\cite{zhang2021motif}, 
which contains 250K unlabeled molecules collected from the ZINC15 database \cite{sterling2015zinc}.
We use the larger version ZINC15 with 2 million molecules
for methods other than MGSSL~\cite{zhang2021motif}.

\subsubsection{GNN fine-tuning tasks and datasets $\mathcal{D}_{ft}$ }
\label{sssec: exp_setting_data}

%\begin{itemize}[leftmargin=*]
%	\item Pre-training:
%	We follow literatures to utilize 2 million 
%	%	(except for MGSSL, where we use 250K to align with their original implementation) 
%	unlabeled molecules collected from the ZINC15 database \cite{sterling2015zinc} to perform GNN pre-training.

%%	\item Downstream fine-tuning:
%	We follow related works in GNN P\&F to target on the graph-level task: molecular property prediction (MPP) \cite{kearnes2016molecular} as downstream task to evaluate the fine-tuning ability of S2PGNN.
%	MPP is an important and widely explored topic in a variety of domains such as physics, chemistry, and materials science.
%	Specifically, a molecule can be modeled as a graph, where nodes denote atoms, edges denote chemical bonds, and the label is related to molecular toxicity or enzyme binding.
%	Given a series of molecular graphs $\mathcal{G}= \{G_1, \dots, G_N\} =\mathcal{G}_{L} \cup \mathcal{G}_{U}$, only a subset of molecules $\mathcal{G}_{L} \subset \mathcal{G}$ with corresponding properties $\mathcal{P}_{L}$ are known. The aim of MPP is to 
%	%train the model with labeled $(\mathcal{G}_{L}, \mathcal{P}_{L})$ such that it can be leveraged to infer 
%	learn graph representations $\mathbf{H}_{G}$ ($\forall G \in \mathcal{G}$) that can be leveraged to infer properties $\mathcal{P}_{U}$ for those unlabeled molecules $\mathcal{G}_{U} \subset \mathcal{G}$.
%	%To thoroughly evaluate S2PGNN's capacity on downstream tasks, 

%{\color{blue}
The proposed S2PGNN is
%our fine-tuning method is 
built on top of existing pre-trained GNNs.
Thus, we follow the vast majority of existing pre-trained GNNs to focus on the graph-level tasks (including graph classification and graph regression) for downstream evaluation.
%}
%Therefore, in this paper, 
Specifically, 
we mainly conduct graph-level experiments on 
downstream molecular property prediction (MPP),
which 
is an important task for a variety of domains (e.g., physics, chemistry, and materials science). 
In MPP, a molecule is represented as a graph, where nodes and edges denote atoms and bonds, and labels are related to molecular toxicity or enzyme binding properties. The aim of MPP is to predict the properties for unlabeled molecules.
%\textcolor{brown}
%{
%	MPP \cite{kearnes2016molecular} is an important and widely explored topic in a variety of domains such as physics, chemistry, and materials science.
%	Specifically, a molecule can be modeled as a graph, where nodes denote atoms, edges denote chemical bonds, and the label is related to molecular toxicity or enzyme binding.
%	Given a series of molecular graphs $\mathcal{G}= \{G_1, \dots, G_N\} =\mathcal{G}_{L} \cup \mathcal{G}_{U}$, only a subset of molecules $\mathcal{G}_{L} \subset \mathcal{G}$ with corresponding properties $\mathcal{P}_{L}$ are known. The aim of MPP is to 
%	%train the model with labeled $(\mathcal{G}_{L}, \mathcal{P}_{L})$ such that it can be leveraged to infer 
%	learn graph representations $\mathbf{H}_{G}$ ($\forall G \in \mathcal{G}$) that can be leveraged to infer properties $\mathcal{P}_{U}$ for those unlabeled molecules $\mathcal{G}_{U} \subset \mathcal{G}$.}
We employ ROC-AUC and RMSE for evaluating the classific and regressive MPP tasks, respectively.
For datasets with multiple prediction tasks (see Tab. \ref{tab: exp_data}), we report average results over all their tasks.

\begin{table}[!t]
	\caption{Summary of base GNN pre-training methods.}
	\label{tab: exp_pretrain_obj}
	\centering
	\vspace{-5px}
	\setlength\tabcolsep{4pt}
	\begin{tabular}{l|l|l}
		\hline \hline
		\bf Method & \bf SSL Strategy $\mathcal{L}_{ssl}(\cdot)$ &\bf  SSL Data $\mathcal{D}_{ssl}$
		\\ \hline \hline
		Infomax & Contrastive Learning (CL) & ZINC15 (2M)
		\\ \hline
		EdgePred & Autoencoding (AE) &ZINC15 (2M)
		\\ \hline
		ContextPred & Context Prediction (CP) &ZINC15 (2M)
		\\ \hline 
		AttrMasking & Masked Component Modeling (MCM) &ZINC15 (2M)
		\\ \hline
		GraphCL & Contrastive Learning (CL) &ZINC15 (2M)
		\\ \hline
		GraphLoG & Contrastive Learning (CL) &ZINC15 (2M)
		\\ \hline
		MGSSL & Autoregressive Modeling (AM) &ZINC15 (250K)
		\\ \hline
		SimGRACE & Contrastive Learning (CL) &ZINC15 (2M)
		\\ \hline
		GraphMAE  & AutoEncoding (AE) &ZINC15 (2M)
		\\ \hline 
		Mole-BERT & Masked Component Modeling (MCM) &ZINC15 (2M)
		\\ \hline
		\hline
	\end{tabular}
	\vspace{-5px}
\end{table}

As shown in Tab. \ref{tab: exp_b1},
we follow the related literature 
%\textcolor{teal}{
\cite{hu2019strategies, rong2020self}
to adopt 8 popular benchmark datasets: BBBP \cite{martins2012bayesian}, Tox21 \cite{tox212014}, ToxCast \cite{richard2016toxcast}, SIDER \cite{kuhn2016sider}, ClinTox \cite{novick2013sweetlead}, BACE \cite{subramanian2016computational}, ESOL \cite{delaney2004esol}, and Lipophilicity (Lipo) \cite{gaulton2012chembl}
%}
that are provided in MoleculeNet \cite{wu2018moleculenet}.
%\footnote{
%	+++ \# shimin: better add citation after every data if any
%}
The selected datasets are from several domains,
including pharmacology, biophysics, and physical chemistry. 
%pharmacology (BBBP, Tox21, ToxCast, SIDER, and ClinTox), biophysics (BACE), and physical chemistry (ESOL, and Lipophilicity (Lipo)). 
%6 of them (except for ESOL and Lipo) are used for binary molecular property prediction task and the rest 2 serve for regression task to be more comprehensive. 
Among them,
ESOL and Lipo are used for graph regression, while the rest are for graph classification. 
%The adopted downstream datasets all fall within the low-data regime and 
As for data split, 
we utilize \textit{scaffold-split} \cite{ramsundar2019deep} to split molecular datasets according to their substructures as suggested by \cite{hu2019strategies, rong2020self}.
It provides a more challenging yet more realistic split to deal with real-world applications (where out-of-distribution predictions are often required) compared with random-split.
Note that AutoGNNs are excluded for empirical comparisons because they focus on searching the GNN architecture (as in Eq.~\eqref{eq: autognn_training}), which thereby 
cannot be extended to solve the problem of searching fine-tuning strategies in ours (as in Definition \ref{def: problem} and Eq.~\eqref{eq: obj_1}).

\begin{table*}[!t]
	\caption{The performance comparison between proposed S2PGNN and vanilla fine-tuning (see Sec. \ref{ssec: related_pf})
		with different pre-training objectives (see Sec. \ref{ssec: exp_b}) and fixed GIN backbone.
		The rightmost column of each task type averages S2PGNN's gain/reduction over vanilla fine-tuning across all involved datasets given the specific pre-training objectives.}
	\label{tab: exp_b1}
	\centering
	\setlength\tabcolsep{5.5pt}
	\begin{tabular}{c|cccccc|cc|c}
		\hline
		\hline
		&
		\multicolumn{6}{c|}{\bf Classification (ROC-AUC (\%)) $\uparrow$}
		&
		\multicolumn{2}{c|}{\bf Regression (RMSE) $\downarrow$}
		&\multirow{2}{*}{\bf \makecell{Avg. \\ Gain}}
		\\
		\cline{1-9}
		\bf Dataset
		&BBBP 
		&Tox21 
		&ToxCast 
		&SIDER 
		&ClinTox
		&BACE 
		%		&\makecell{Avg. $\uparrow$}
		&ESOL 
		&Lipo
		&
		\\
		\hline
		Infomax~\cite{velickovic2019deep}
		&68.4 $\pm$ 1.7
		&75.6 $\pm$ 0.5
		&62.5 $\pm$ 0.8
		&58.3 $\pm$ 0.7
		&71.3 $\pm$ 2.6
		&75.5 $\pm$ 2.3
		%		&
		&2.6 $\pm$ 0.1
		&1.0 $\pm$ 0.1
		&\multirow{2}{*}{$+17.7\%$}
		\\
		Infomax + S2PGNN
		&\bf 69.9 $\pm$ 1.4
		&\bf76.7 $\pm$ 0.5
		&\bf65.8 $\pm$ 0.4
		&\bf62.3 $\pm$ 1.3
		&\bf74.8 $\pm$ 3.8
		&\bf82.3 $\pm$ 1.3
		%		&
		&\bf1.5 $\pm$ 0.3
		&\bf0.8 $\pm$ 0.0
		&
		\\
		\hline
		EdgePred~\cite{hamilton2017inductive}
		&67.2 $\pm$ 2.9
		&75.8 $\pm$ 0.9
		&63.9 $\pm$ 0.4
		&60.5 $\pm$ 0.8
		&65.7 $\pm$ 4.1
		&79.4 $\pm$ 1.4
		%		&
		&2.8 $\pm$ 0.0
		&1.0 $\pm$ 0.1
		&\multirow{2}{*}{$+14.4\%$}
		\\
		EdgePred + S2PGNN
		&\bf69.1 $\pm$ 0.8
		&\bf77.1 $\pm$ 0.8
		&\bf66.2 $\pm$ 0.3
		&\bf62.3 $\pm$ 0.5
		&\bf71.9 $\pm$ 1.1
		&\bf82.2 $\pm$ 1.1
		%		&
		&\bf1.7 $\pm$ 0.2
		&\bf0.9 $\pm$ 0.0
		&
		\\
		\hline
		ContextPred~\cite{hu2019strategies}
		&69.0 $\pm$ 0.9
		&76.0 $\pm$ 0.4
		&63.5 $\pm$ 0.4
		&60.7 $\pm$ 0.6
		&69.7 $\pm$ 1.4
		&80.6 $\pm$ 0.8
		%		&
		&2.8 $\pm$ 0.4
		&1.1 $\pm$ 0.0
		&\multirow{2}{*}{$+15.1\%$}
		\\
		ContextPred + S2PGNN
		&\bf70.9 $\pm$ 1.3
		&\bf76.3 $\pm$ 0.4
		&\bf67.0 $\pm$ 0.5
		&\bf62.8 $\pm$ 0.3
		&\bf75.9 $\pm$ 2.2
		&\bf82.6 $\pm$ 0.7
		%		&
		&\bf1.7 $\pm$ 0.2
		&\bf0.9 $\pm$ 0.0
		&
		\\
		\hline
		AttrMasking~\cite{hu2019strategies}
		&65.1 $\pm$ 2.3
		&76.7 $\pm$ 0.6
		&64.4 $\pm$ 0.3
		&60.6 $\pm$ 0.9
		&72.0 $\pm$ 4.2
		&79.5 $\pm$ 0.7
		%		&
		&2.8 $\pm$ 0.1
		&1.1 $\pm$ 0.0
		&\multirow{2}{*}{$+15.6\%$}
		\\
		AttrMasking + S2PGNN
		&\bf71.9 $\pm$ 1.1
		&\bf77.3 $\pm$ 0.4
		&\bf66.8 $\pm$ 0.5
		&\bf62.9 $\pm$ 0.4
		&\bf74.8 $\pm$ 3.1
		&\bf82.7 $\pm$ 0.8
		%		&
		&\bf1.7 $\pm$ 0.1
		&\bf0.9 $\pm$ 0.0
		&
		\\
		\hline
		GraphCL~\cite{you2020graph} 
		&68.3 $\pm$ 1.6
		&74.1 $\pm$ 0.8
		&62.6 $\pm$ 0.5
		&59.7 $\pm$ 1.1
		&71.4 $\pm$ 6.2
		&75.8 $\pm$ 2.7
		%		&
		&2.5 $\pm$ 0.1
		&1.0 $\pm$ 0.0
		&\multirow{2}{*}{$+10.1\%$}
		\\
		GraphCL + S2PGNN
		&\bf70.8 $\pm$ 1.1
		&\bf76.8 $\pm$ 0.5
		&\bf66.6 $\pm$ 0.3
		&\bf62.4 $\pm$ 1.2
		&\bf75.2 $\pm$ 3.4
		&\bf82.6 $\pm$ 2.3
		%		&
		&\bf1.9 $\pm$ 0.1
		&\bf0.9 $\pm$ 0.0
		&
		\\
		\hline
		GraphLoG~\cite{xu2021self}
		&66.5 $\pm$ 2.3
		&75.3 $\pm$ 0.3
		&63.3 $\pm$ 0.5
		&57.8 $\pm$ 1.2
		&69.6 $\pm$ 5.8
		&80.1 $\pm$ 2.4
		%		&
		&2.5 $\pm$ 0.1
		&1.0 $\pm$ 0.0
		&\multirow{2}{*}{$+9.1\%$}
		\\
		GraphLoG + S2PGNN
		&\bf69.9 $\pm$ 1.5
		&\bf76.8 $\pm$ 0.3
		&\bf66.1 $\pm$ 0.2
		&\bf62.0 $\pm$ 1.0
		&\bf75.8 $\pm$ 2.1
		&\bf85.1 $\pm$ 1.3
		%		&
		&\bf2.0 $\pm$ 0.1
		&\bf0.9 $\pm$ 0.0
		&
		\\
		\hline
		MGSSL~\cite{zhang2021motif}
		&67.4 $\pm$ 1.8
		&75.0 $\pm$ 0.6
		&63.1 $\pm$ 0.5
		&58.0 $\pm$ 1.2
		&68.1 $\pm$ 5.4
		&81.2 $\pm$ 3.3
		%		&
		&2.4 $\pm$ 0.1
		&1.0 $\pm$ 0.0
		&\multirow{2}{*}{$+9.6\%$}
		\\
		MGSSL + S2PGNN
		&\bf69.4 $\pm$ 1.8
		&\bf77.0 $\pm$ 0.6
		&\bf66.3 $\pm$ 0.4
		&\bf62.8 $\pm$ 1.2
		&\bf76.7 $\pm$ 2.4
		&\bf85.2 $\pm$ 1.2
		%		&
		&\bf1.9 $\pm$ 0.2
		&\bf0.9 $\pm$ 0.0
		&
		\\
		\hline
		SimGRACE~\cite{xia2022simgrace}
		&67.9 $\pm$ 0.6
		&73.9 $\pm$ 0.5
		&61.9 $\pm$ 0.5
		&59.1 $\pm$ 0.7
		&61.1 $\pm$ 3.8
		&75.5 $\pm$ 1.4
		%		&
		&2.6 $\pm$ 0.1
		&1.0 $\pm$ 0.0
		&\multirow{2}{*}{$+16.5\%$}
		\\
		SimGRACE + S2PGNN
		&\bf69.3 $\pm$ 0.9
		&\bf75.9 $\pm$ 0.2
		&\bf65.8 $\pm$ 0.3
		&\bf62.3 $\pm$ 0.6
		&\bf73.6 $\pm$ 3.2
		&\bf83.9 $\pm$ 1.5
		%		&
		&\bf1.6 $\pm$ 0.3
		&\bf0.9 $\pm$ 0.0
		&
		\\
		\hline
		GraphMAE~\cite{hou2022graphmae}
		&70.0 $\pm$ 1.0
		&75.1 $\pm$ 1.4
		&64.4 $\pm$ 2.0
		&60.7 $\pm$ 1.1
		&71.7 $\pm$ 7.2
		&79.8 $\pm$ 4.2
		%		&
		&2.3 $\pm$ 0.4
		&1.0 $\pm$ 0.1
		&\multirow{2}{*}{$+10.3\%$}
		\\
		GraphMAE + S2PGNN
		&\bf70.3 $\pm$ 0.8
		&\bf76.7 $\pm$ 0.8
		&\bf66.4 $\pm$ 0.5
		&\bf62.2 $\pm$ 0.6
		&\bf77.0 $\pm$ 2.8
		&\bf82.2 $\pm$ 1.2
		%		&
		&\bf1.6 $\pm$ 0.2
		&\bf0.9 $\pm$ 0.0
		&
		\\
		\hline
		Mole-BERT~\cite{xia2022mole} 
		&70.6 $\pm$ 1.4
		&77.4 $\pm$ 2.2
		&65.4 $\pm$ 1.9
		&61.9 $\pm$ 2.2
		&75.6 $\pm$ 3.1
		&77.4 $\pm$ 4.2
		%		&
		&2.4 $\pm$ 0.4
		&1.0 $\pm$ 0.1
		&\multirow{2}{*}{$+14.5\%$}
		\\
		Mole-BERT + S2PGNN
		&\bf71.4 $\pm$ 0.4
		&\bf79.5 $\pm$ 0.4
		&\bf67.8 $\pm$ 0.2
		&\bf63.8 $\pm$ 0.5
		&\bf76.5 $\pm$ 0.5
		&\bf84.2 $\pm$ 0.7
		%		&
		&\bf1.5 $\pm$ 0.2
		&\bf0.8 $\pm$ 0.0
		&
		\\
		\hline
		\hline
	\end{tabular}
\end{table*}

\subsubsection{Implementation details}
\label{sssec: exp_setting_details}
For GNN pre-training methods, we employ their officially released pre-trained models to conduct the next stage fine-tuning.
Readers may refer to the original papers for their detailed pre-training settings.

S2PGNN and other fine-tuning baselines are evaluated with the same protocol for rigorously fair comparisons.
We follow pioneering literature \cite{hu2019strategies} to setup fine-tuning configurations.
Specifically,
we leverage the simple linear classifier as downstream prediction head.
To fine-tune the whole model, 
we use Adam optimizer with a learning rate of 1e-3. We set batch size as 32 and dropout rate as 50\%. We perform fine-tuning for 100 epochs with early stopping based on validation set. 
The ratio to split train/validation/test set is 80\%/10\%/10\%. 
We run all experiments for 10 times with different random seeds and report the mean results (standard deviations).

\subsection{The implementation S2PGNN with Pre-trained GNNs and comparison with vanilla fine-tuning}
\label{ssec: exp_b}

%Despite the blooming development of GNN pre-training methods,
%the specific strategies for GNN fine-tuning (Eq. \eqref{eq: related_pf_2}) is still scarce. 
As discussed in Sec. \ref{ssec: related_pf}, vanilla fine-tuning is still the most prevalent way to leverage pre-trained GNNs in existing works. 
Thus, we first implement and compare the gains of S2PGNN over vanilla strategy on top of 10 classic GNN pre-training methods (Tab. \ref{tab: exp_pretrain_obj}) from various categories:
%that cover the mainstream of existing GNN pre-training methods as summarized in the following.

%{\color{brown}
		\begin{itemize}[leftmargin=*]
			\item AutoEncoding (AE): 
			Given the partial access to graph, AE methods
%			(e.g., EdgePred \cite{hamilton2017inductive} and GraphMAE \cite{hou2022graphmae})
			propose to reconstruct the input graph via autoencoder architecture \cite{hinton2006reducing}.
			Let $\tilde{G}$ be reconstructed graph,
			then their objective is:
			$\mathcal{L}_{ssl}(\cdot) = -\sum_{G \in \mathcal{D}_{ssl}} \log p(\tilde{G}|G)$.
%			\begin{equation}
%				\mathcal{L}_{ssl}(\cdot) = -\sum_{G \in \mathcal{D}_{ssl}} \log p(\tilde{G}|G).
%				\label{eq: ae}
%			\end{equation}
			
			\item Autoregressive Modeling (AM): 
			AM methods 
%			(e.g., MGSSL~\cite{zhang2021motif})
			factorize input graph 
			$G$
			as a sequence of components 
			$\mathcal{C} =\{C_1, C_2, \dots\}$
			(e.g., nodes, edges, and subgraphs) 
			with some preset ordering and perform graph reconstruction in an autoregressive manner:
			$\mathcal{L}_{ssl}(\cdot) = -\sum_{G \in \mathcal{D}_{ssl}}\sum_{i=1}^{|\mathcal{C}|} \log p ({C}_{i}| {C}_{\textless i})$.
%			\begin{equation}
%				\mathcal{L}_{ssl}(\cdot) = -\sum_{G \in \mathcal{D}_{ssl}}\sum_{i=1}^{|\mathcal{C}|} \log p ({C}_{i}| {C}_{\textless i}).
%				\label{eq: am}
%			\end{equation}

			\item Masked Component Modeling (MCM):
			MCM works 
%			(e.g., AttrMasking \cite{hu2019strategies} and Mole-BERT \cite{xia2022mole})
			masks out some components of input graphs (e.g., nodes, edges, and subgraphs), 
			then aim to recover those masked ones $m(G)$
			through the remaining ones $G \backslash m(G)$:
			$\mathcal{L}_{ssl}(\cdot) = -\sum_{G \in \mathcal{D}_{ssl}}\sum_{\hat{G} \in m(G)} \log p (\hat{G}| G \backslash m(G))$.
%			\begin{equation}
%				\mathcal{L}_{ssl}(\cdot) = -\sum_{G \in \mathcal{D}_{ssl}}\sum_{\hat{G} \in m(G)} \log p (\hat{G}| G \backslash m(G)).
%				\label{eq: mcm}
%			\end{equation}
			
			\item Context Prediction (CP): 
			CP explores graph structures and uses contextual information to design pre-training objectives. 
			Let $t=1$ if subgraph ${C}_{1}$ and surrounding context ${C}_{2}$ share the same center node, otherwise $t=0$. 
			CP leverages subgraphs to predict their surrounding context structures:
			$\mathcal{L}_{ssl}(\cdot) = -\sum_{G \in \mathcal{D}_{ssl}} \log p(t|{C}_{1}, {C}_{2})$.
%			\begin{equation}
%				\mathcal{L}_{ssl}(\cdot) = -\sum_{G \in \mathcal{D}_{ssl}} \log p(t|{C}_{1}, {C}_{2}).
%				\label{eq: cp}
%			\end{equation}
			
			\item 
			Contrastive Learning (CL): 
			CL conducts pre-training via maximizing the agreement between a pair of similar inputs,
			including Cross-Scale Contrastive Learning 
			and Same-Scale Contrastive Learning.
			The former 
%			(e.g., Infomax \cite{velickovic2019deep}) 
			contrasts a pair of graph and its local substructure $(G, C)$ against negative pairs $(G, C^{-})$:
			$\mathcal{L}_{ssl}(\cdot)\! =\! -\!\sum_{G \in \mathcal{D}_{ssl}}\![\log s(G, C) \!-\! \sum_{{C}^{-}} \log s(G, {C}^{-})]]$,
%			\begin{equation}
%				\mathcal{L}_{ssl}(\cdot)\! =\! -\!\sum_{G \in \mathcal{D}_{ssl}}\![\log s(G, C) \!-\! \sum_{{C}^{-}} \log s(G, {C}^{-})]],
%				\label{eq: cs}
%			\end{equation}
			where $s(\cdot, \cdot)$ is the similarity function. 
			The latter
%			(e.g., 
%			GraphCL \cite{you2020graph},
%			SimGRACE \cite{xia2022simgrace}, 
%			and GraphLoG \cite{xu2021self})
			maximizes the agreement between the augmented graph and its anchor graph $(G, G^{+})$ and meanwhile repel negative pairs $(G, G^{-})$:
			$\mathcal{L}_{ssl}(\cdot)\! =\! -\!\sum_{G \in \mathcal{D}_{ssl}}\![\log s(G, G^{+}) \!-\! \sum_{{C}^{-}} \log s(G, G^{-})]]$.
%			\begin{equation}
%				\mathcal{L}_{ssl}(\cdot)\! =\! -\!\sum_{G \in \mathcal{D}_{ssl}}\![\log s(G, G^{+}) \!-\! \sum_{{C}^{-}} \log s(G, G^{-})]].
%				\label{eq: ss}
%			\end{equation}
		\end{itemize}

%}

The main results on 8 downstream datasets are reported in Tab. \ref{tab: exp_b1}.
When equipped with S2PGNN, the pre-trained GNNs 
consistently
demonstrate
better fine-tuning performances on 
the
graph classification and graph regression tasks than the vanilla fine-tuning.
The gains are consistent on different datasets with diverse characteristics (see Tab. \ref{tab: exp_data}), and the average improvement across all datasets is significant ($9.1\% \sim 17.7\%$). 
%S2PGNN demonstrates superior results than the standard fine-tuning strategy across various datasets.
Moreover, we also observe the superiority of S2PGNN is agnostic to GNN pre-training configurations, such as SSL strategy, SSL data, and attained pre-trained models (see Sec. \ref{sssec: exp_setting_base} and Tab. \ref{tab: exp_pretrain_obj}). 
To summarize, observations from Tab. \ref{tab: exp_b1} validate that S2PGNN fine-tuning provides a promisingly better solution than the vanilla strategy to achieve the better utilization of various pre-trained GNNs.

%\footnote{\# zhili: for Table~\ref{tab: exp_b1} and others, maybe we show relative improvements on all 8 datasets (instead of current way to separate them)}

%\footnote{\# zhili: for results analysis, may refer to 
%	1) "SpotTune: Transfer Learning through Adaptive Fine-tuning", 
%	2) "Raise a Child in Large Language Model: Towards Effective and Generalizable Fine-tuning", 
%	3) "Fine-Tuning Graph Neural Networks via Graph Topology induced Optimal Transport", 
%	and "7898 submission"}
%}

\begin{table*}[!t]
	\caption{The performance comparison between proposed S2PGNN fine-tuning and other fine-tuning strategies
		with fixed ContextPred pre-training objective and GIN backbone architecture.}
	\label{tab: exp_c1}
	\centering
	%	\vspace{-10px}
	\setlength\tabcolsep{12pt}
	%\scalebox{1}{
	\begin{tabular}{c|cccccc|c}
		\hline \hline
		&
		\multicolumn{6}{c|}{\bf Classification (ROC-AUC (\%)) $\uparrow$}
		&\multirow{2}{*}{\bf Avg.}
		\\
		\cline{1-7}
		\bf Dataset
		&BBBP 
		&Tox21 
		&ToxCast 
		&SIDER 
		&ClinTox
		&BACE 
		&
		\\
		\hline
		Vanilla fine-tuning
		&68.0 $\pm$ 2.0
		&75.7 $\pm$ 0.7
		&63.9 $\pm$ 0.6
		&60.9 $\pm$ 0.6
		&65.9 $\pm$ 3.8
		&79.6 $\pm$ 1.2
		&69.0
		\\
		\hline
		\textit{$L^2$-SP}~\cite{xuhong2018explicit}
		&68.2 $\pm$ 0.7
		&73.6 $\pm$ 0.8
		&62.4 $\pm$ 0.3 
		&61.1 $\pm$ 0.7
		&68.1 $\pm$ 3.7
		&82.2 $\pm$ 2.4
		&69.3
		\\
		DELTA~\cite{li2019delta}
		&67.8 $\pm$ 0.8
		&75.2 $\pm$ 0.5
		&63.3 $\pm$ 0.5
		&62.2 $\pm$ 0.4
		&73.4 $\pm$ 3.0
		&81.8 $\pm$ 1.1
		&70.6
		\\
		BSS~\cite{chen2019catastrophic}
		&68.1 $\pm$ 1.4
		&75.9 $\pm$ 0.8
		&63.9 $\pm$ 0.4
		&60.9 $\pm$ 0.8
		&70.9 $\pm$ 5.1
		&82.4 $\pm$ 1.8
		&70.4
		\\
		StochNorm~\cite{kou2020stochastic}
		&69.3 $\pm$ 1.6
		&74.9 $\pm$ 0.6
		&63.4 $\pm$ 0.5
		&61.0 $\pm$ 1.1
		&65.5 $\pm$ 4.2
		&80.5 $\pm$ 2.7
		&69.1
		\\
		\hline
		GTOT-tuning~\cite{zhang2022fine} 
		&70.0 $\pm$ 1.7
		&75.2 $\pm$ 0.9
		&63.0 $\pm$ 0.5
		&{63.1 $\pm$ 0.6}
		&71.8 $\pm$ 5.4
		&82.6 $\pm$ 2.0
		&71.0
		\\
		\hline
		S2PGNN (ContextPred)
		%		&\textbf{71.0 $\pm$ 0.3}
		%		&\textbf{75.7 $\pm$ 0.1}
		%		&\textbf{66.6 $\pm$ 0.1}
		%		&\underline{63.1 $\pm$ 0.8}
		%		&\textbf{78.1 $\pm$ 2.2}
		%		&\underline{83.0 $\pm$ 1.3}
		%		&
		&{70.9 $\pm$ 1.3} 
		&{76.3 $\pm$ 0.4} 
		&{67.0 $\pm$ 0.5} 
		&62.8 $\pm$ 0.3
		&{75.9 $\pm$ 2.2} 
		&{82.6 $\pm$ 0.7} 
		&$72.6$
		\\
		%		\color{blue}
		S2PGNN (Mole-BERT)
		%		&\color{blue} \bf71.4 $\pm$ 0.4
		%		&\color{blue} \bf79.5 $\pm$ 0.4
		%		&\color{blue} \bf67.8 $\pm$ 0.2
		%		&\color{blue} \bf63.8 $\pm$ 0.5
		%		&\color{blue} \bf76.5 $\pm$ 0.5
		%		&\color{blue} \bf84.2 $\pm$ 0.7
		%		&\color{blue} \bf73.9
		&\bf71.4 $\pm$ 0.4
		&\bf79.5 $\pm$ 0.4
		&\bf67.8 $\pm$ 0.2
		&\bf63.8 $\pm$ 0.5
		&\bf76.5 $\pm$ 0.5
		&\bf84.2 $\pm$ 0.7
		&\bf73.9
		\\
		\hline
		\hline
	\end{tabular}
	%	\vspace{-10px}
\end{table*}

\begin{table*}[!t]
	\caption{The performance of more
		fine-tuning strategies that are excluded from S2PGNNX's design space
		with fixed ContextPred pre-training objective and GIN backbone architecture.
		Second best results are marked with \underline{underline}.
	}
	\label{tab: exp_c2}
	\centering
	\setlength\tabcolsep{13.5pt}
	\begin{tabular}{c|cccccc|c}
		\hline \hline
		&
		\multicolumn{6}{c|}{\bf Classification (ROC-AUC (\%)) $\uparrow$}
		&\multirow{2}{*}{\bf Avg.}
		\\
		\cline{1-7}
		\bf Dataset
		&BBBP 
		&Tox21 
		&ToxCast 
		&SIDER 
		&ClinTox
		&BACE 
		&
		\\
		\hline
		Vanilla tuning
		&68.0 $\pm$ 2.0
		&\underline{75.7 $\pm$ 0.7}
		&63.9 $\pm$ 0.6
		&60.9 $\pm$ 0.6
		&\underline{65.9 $\pm$ 3.8}
		&79.6 $\pm$ 1.2
		&\underline{69.0}
		\\
		\hline
		%		Last-$k$ ($k=4$)
		%		&68.4 $\pm$ 1.5
		%		&75.5 $\pm$ 0.5
		%		&63.8 $\pm$ 0.4	
		%		&61.2 $\pm$ 0.7
		%		&68.5 $\pm$ 3.0
		%		&79.7 $\pm$ 1.4
		%		&
		%		\\
		Feature extractor
		&58.9 $\pm$ 0.4
		&68.7 $\pm$ 0.4
		&59.3 $\pm$ 0.2
		&59.9 $\pm$ 0.3
		&40.5 $\pm$ 2.7
		&61.6 $\pm$ 4.5
		&58.2
		\\
		\hline
		Last-$k$ ($k=3$)
		&\underline{68.1 $\pm$ 0.9}
		&75.1 $\pm$ 0.4
		&\underline{64.0 $\pm$ 0.5}
		&60.7 $\pm$ 0.5
		&63.9 $\pm$ 4.5
		&79.1 $\pm$ 1.2
		&68.5
		\\
		Last-$k$ ($k=2$)
		&65.3 $\pm$ 1.0
		&74.5 $\pm$ 0.5
		&63.0 $\pm$ 0.7
		&\underline{61.6 $\pm$ 0.6}
		&64.0 $\pm$ 3.5
		&\underline{80.6 $\pm$ 1.2}
		&68.2
		\\
		Last-$k$ ($k=1$)
		&64.6 $\pm$ 1.3
		&73.0 $\pm$ 0.5
		&61.4 $\pm$ 0.5
		&60.8 $\pm$ 0.5
		&66.1 $\pm$ 2.4
		&76.6 $\pm$ 0.8
		&67.1
		\\
		\hline
		Adapter ($m=2$)
		&61.2 $\pm$ 0.4
		&71.4 $\pm$ 0.3
		&60.6 $\pm$ 0.1
		&59.6 $\pm$ 0.3
		&45.3 $\pm$ 3.1
		&71.1 $\pm$ 0.9
		&61.6
		\\
		Adapter ($m=4$)
		&62.5 $\pm$ 0.7
		&71.7 $\pm$ 0.3
		&60.6 $\pm$ 0.2
		&59.6 $\pm$ 0.3
		&47.4 $\pm$ 2.2
		&74.4 $\pm$ 1.8
		&62.7
		\\
		Adapter ($m=8$)
		&63.9 $\pm$ 0.7
		&71.8 $\pm$ 0.3
		&60.5 $\pm$ 0.4
		&59.9 $\pm$ 0.3
		&50.0 $\pm$ 1.0
		&76.8 $\pm$ 0.9
		&63.8
		\\
		\hline
		%		\textit{$L^2$-SP}
		%		&68.2 $\pm$ 0.7
		%		&73.6 $\pm$ 0.8
		%		&62.4 $\pm$ 0.3 
		%		&61.1 $\pm$ 0.7
		%		&68.1 $\pm$ 3.7
		%		&82.2 $\pm$ 2.4
		%		&
		%		\\
		%		DELTA
		%		&67.8 $\pm$ 0.8
		%		&75.2 $\pm$ 0.5
		%		&63.3 $\pm$ 0.5
		%		&62.2 $\pm$ 0.4
		%		&73.4 $\pm$ 3.0
		%		&81.8 $\pm$ 1.1
		%		&
		%		\\
		%		BSS
		%		&68.1 $\pm$ 1.4
		%		&75.9 $\pm$ 0.8
		%		&63.9 $\pm$ 0.4
		%		&60.9 $\pm$ 0.8
		%		&70.9 $\pm$ 5.1
		%		&82.4 $\pm$ 1.8
		%		&
		%		\\
		%		SotchNorm 
		%		&69.3 $\pm$ 1.6
		%		&74.9 $\pm$ 0.6
		%		&63.4 $\pm$ 0.5
		%		&61.0 $\pm$ 1.1
		%		&65.5 $\pm$ 4.2
		%		&80.5 $\pm$ 2.7
		%		&
		%		\\
		%		\hline
		%		GTOT-tuning 
		%		&70.0 $\pm$ 2.3
		%		&75.6 $\pm$ 0.7
		%		&64.0 $\pm$ 0.3
		%		&\textbf{63.5 $\pm$ 0.6}
		%		&72.0 $\pm$ 5.4
		%		&\textbf{83.4 $\pm$ 1.9}
		%		&
		%		\\
		%		\hline
		S2PGNN 
		%		&\textbf{71.0 $\pm$ 0.3}
		%		&\textbf{75.7 $\pm$ 0.1}
		%		&\textbf{66.6 $\pm$ 0.1}
		%		&\underline{63.1 $\pm$ 0.8}
		%		&\textbf{78.1 $\pm$ 2.2}
		%		&\underline{83.0 $\pm$ 1.3}
		%		&
		&\textbf{70.9 $\pm$ 1.3} 
		&\textbf{76.3 $\pm$ 0.4} 
		&\textbf{67.0 $\pm$ 0.5} 
		&\textbf{62.8 $\pm$ 0.3} 
		&\textbf{75.9 $\pm$ 2.2} 
		&\textbf{82.6 $\pm$ 0.7} 
		&\textbf{72.6}
		\\
		\hline
		\hline
	\end{tabular}
\end{table*}

\subsection{Comparison with other fine-tuning strategies}
\label{ssec: exp_c}

%{\color{red}
%+++
%To answer Q2 and Q3,
%we should first compare S2PGNN with existing GNN fine-tuning strategies,
%then show whether those fine-tuning strategies that are not included into our space
%is effective or not
%+++
%}

As discussed in Sec.~\ref{ssec: related_pf},
we first investigate the vanilla fine-tuning 
and regularized fine-tuning strategies since they have already been adapted to the GNN area.
Then,
we explore more
fine-tuning methods on other domains (e.g., computer vision)
but have not been discussed in the GNN community.

\subsubsection{GNN Fine-tuning Baselines}
\label{sssec: exp_setting_baseline}

%Despite the blooming development of GNN pre-training methods,
%the specific strategies for GNN fine-tuning (Eq. \eqref{eq: related_pf_2}) is still scarce. 
%As discussed in Sec. \ref{sssec: related_pf_tune}, standard fine-tuning is probably still the most prevalent strategy to leverage pre-trained GNNs among existing literatures. 
%Thus, we first compare the gains of S2PGNN over standard fine-tuning on top of various GNN pre-training methods.
%{\color{red}
%+++
%analyze Tab.~\ref{tab: exp_b1}
%+++
%}

Apart from the vanilla strategy (see main results in Sec. \ref{ssec: exp_b}), 
we also compare S2PGNN with another GNN fine-tuning work GTOT-Tuning~\cite{zhang2022fine},
which belongs to regularized fine-tuning.
%please explain here why AUX-TS and WordReg are missing
We exclude AUX-TS \cite{han2021adaptive} and WordReg \cite{xia2022towards} for baseline comparisons because: AUX-TS targets on different downstream tasks, and WordReg uses different data split with \cite{hu2019strategies} 
%{\color{teal}
but its code is not publicly available for reproducing results.
%}
%1) AUX-TS focus on the NC and LP as downstream tasks, and datasets are also from different domains; 2) WordReg uses different data split, but its code is not published --> cannot compare these 2 methods
Besides,
we 
additionally report the results of  
several regularized fine-tuning baselines tailored from the computer vision domain (that are originally designed to fine-tune CNNs),
including
\textit{$L^2$-SP} \cite{xuhong2018explicit}, 
DELTA \cite{li2019delta}, 
BSS \cite{chen2019catastrophic}, and StochNorm \cite{kou2020stochastic}. 
\textit{$L^2$-SP} regularizes on model parameters to induce the fine-tuned weights to be close to pre-trained weights.
DELTA imposes regularization on representations via the attention mechanism.
BSS penalizes small eigenvalues of learned representations to suppress untransferable components. 
StochNorm regularizes on the encoder architecture in a dropout-like way.
Please refer to Sec. \ref{ssec: related_pf} and Tab. \ref{tab: related_ft} for more technical discussions.
%\footnote{\# zhili: to add technical details for \textit{$L^2$-SP} \cite{xuhong2018explicit}, 
%	DELTA \cite{li2019delta}, 
%	BSS \cite{chen2019catastrophic}, and StochNorm \cite{kou2020stochastic} here? copy my previous introductions for these methods}

The comparisons on 6 classific MPP datasets have been summarized in Tab. \ref{tab: exp_c1}.
As shown in Tab. \ref{tab: exp_c1}, 
we first observe the non-negligible 
improvement brought by S2PGNN compared with all baselines.
Among baselines,
regularized techniques (DELTA, BSS) from computer vision domain sometimes yield slightly better results than the vanilla fine-tuning strategy.
However, in several cases (\textit{$L^2$-SP}, StochNorm), the performance gains are not significant. This is probably 
%because that the intrinsic differences among different types of data (image and graph).
due to their ignorance of characteristics in graph data (e.g., graph topology) and specific requirements in fine-tuning GNNs. 
By taking the graph topology into consideration, the method GTOT-Tuning, which is designed specifically to fine-tune GNNs, demonstrates higher performances than regularized variants extended from other domains. 
However, even the most competitive baseline GTOT-Tuning is still inferior to S2PGNN in 5 out of 6 datasets, indicating that 
%{\color{teal}
%the architecture adaption in S2PGNN
the design dimensions proposed in S2PGNN (see Sec. \ref{ssec: method_space} and Tab. \ref{tab: related_ft})
%}
may be indispensable and crucial designs towards more effective GNN fine-tuning.
Furthermore, 
we note that various regularized strategies, such as GTOT-Tuning, is 
%complementary
orthogonal to the proposed S2PGNN. Therefore, it may be promising to combine S2PGNN with other advanced regularized methods, i.e., combine an additional regularization term in S2PGNN's loss function Eq.~\eqref{eq: real_obj_1} and \eqref{eq: real_obj_2} to further boost its performances, which we leave as future works.

\subsubsection{Exploration of Other Fine-tuning Strategies}
\label{sssec: exploration}

\begin{figure*}[!t]
	\centering
	\subfloat[
%	\color{red} 
	An example of fine-tuning strategy with feature extractor.]
	{\label{fig: extractor}
		\includegraphics[width=0.9\linewidth]
		{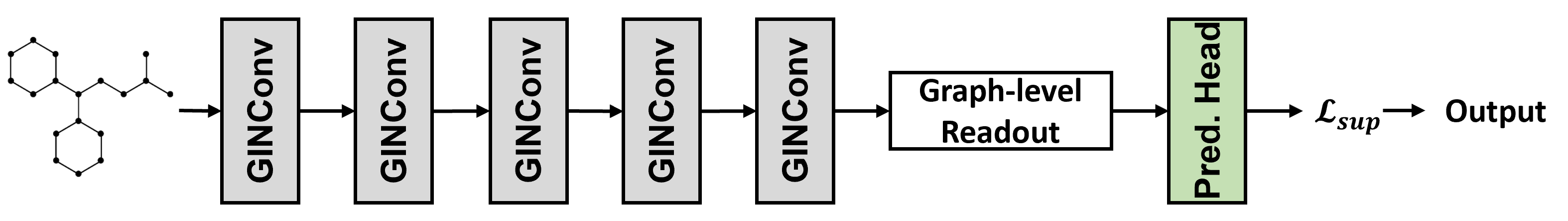}
	}\\
	\subfloat[An example of fine-tuning strategy with last-$k$ ($k=3$) tunable layers.]
	{\label{fig: lastk}
	\includegraphics[width=0.9\linewidth]
	{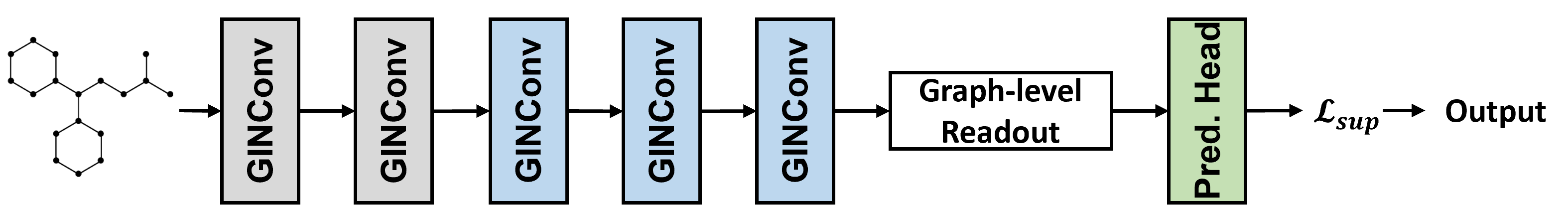}
	}\\
	\subfloat
	[An example of fine-tuning strategy with adapter.
%	($m=5$).
	]
	{\label{fig: adapter}
		\includegraphics[width=1.0\linewidth]
		{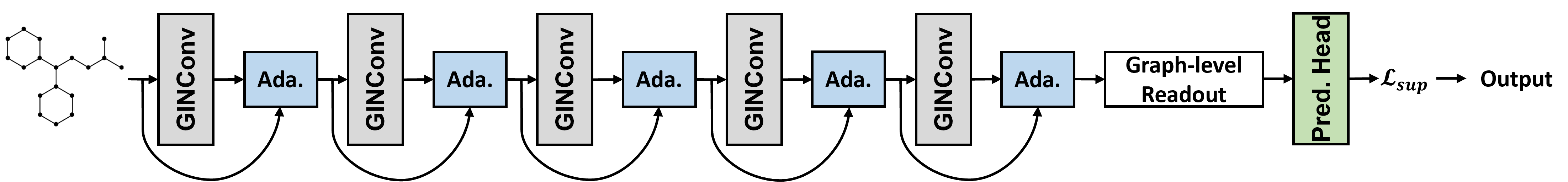}}\\	
	\caption{Illustration to other fine-tuning strategies (refer to Fig. \ref{fig: ours} for the legend).}
	\vspace{-5px}
\end{figure*}

As mentioned in Sec.~\ref{ssec: related_pf} and Remark \ref{remark: ouside_gnn},
fine-tuning has been well explored in the domains beyond GNNs.
Therefore,
we try to investigate the performance of more fine-tuning strategies that are promising in other domains.
However, we conclude that they may not be suitable for fine-tuning GNNs,
thereby we discard them from our search space.
Overall, we explore following fine-tuning strategies:
\begin{itemize}[leftmargin=*]
	
\item Feature Extractor (FE): 
FE \cite{sharif2014cnn} proposes to reuse pre-trained model parameters to achieve better knowledge preservation. Once the pre-training is finished, all pre-trained layers are frozen and pre-trained model works as a pure feature extractor for downstream data. Only a small amount of additional parameters in specific prediction head are further tuned to perform downstream predictions.
%\begin{align}
%&\bm{\theta}^{\ast} = \bm{\theta}_{init},
%\label{eq: fe_1}\\
%&\bm{\omega}^{\ast} = \arg\min_{\bm{\omega}} \mathcal{L}_{ssp} ((g_{\bm{\omega}}(f_{\bm{\theta}_{init}}(\cdot)); \mathcal{D}_{ft}).
%\label{eq: fe_2}
%\end{align}	

\item Last-$k$ Tuning (LKT): With LKT \cite{long2015learning}, only parameters in last-$k$ layers of the pre-trained models are further tuned, while the other initial layers are frozen and keep unchanged. LKT resides between ST and FE, and is popular in the computer version area \cite{long2015learning}. 
In our experiments, we consider $k \in \{1, 2, 3\}$, which means we use around $20\% \sim 60\%$ tunable parameters of the original model (where $k=5$)

\item Adapter-Tuning (AT): 
AT \cite{houlsby2019parameter}
proposes to fine-tune only a small number of extra model parameters to attain the competitive performance, 
which is 
promising
to achieve faster tuning and alleviate the over-fitting issue
in natural language process and computer vision areas.
More specifically,
it adds small and task-specific neural network modules, called adapters, 
to pre-trained models.
These adapters involve feature transformation $\mathbb{R}^{d} \rightarrow \mathbb{R}^{m} \rightarrow \mathbb{R}^{d}$ via the bottleneck architecture, where $m \ll d$ is to ensure  parameter-efficient.
Adapters are inserted between pre-trained layers and
trained on the specific task, 
while the pre-trained layers remain fixed and unchanged to preserve the knowledge learned during pre-training.
In our empirical explorations, we tailor the adapter design in \cite{houlsby2019parameter} and consider adapter size $m \in \{2, 4, 8\}$ to use only around $1.3\% \sim 5.2\%$ tunable parameters of the original model (where $d=300$). 
%\begin{align}
%&\bm{\theta}^{\ast} = \bm{\theta}_{init},
%\label{eq: pet_1}\\
%&\bm{\Theta}^{\ast}, \bm{\omega}^{\ast} = \arg\min_{\bm{\Theta}, \bm{\omega}} \mathcal{L}_{ssp} ((g_{\bm{\omega}}(f_{(\bm{\theta}_{init}, \bm{\Theta})}(\cdot)); \mathcal{D}_{ft}).
%\label{eq: pet_2}
%\end{align}

\end{itemize}

The results on classific MPP tasks are summarized in Tab.~\ref{tab: exp_c2}.
Clearly, S2PGNN demonstrates consistent superior results than other approaches. 
Directly using pre-trained GNN as pure feature extractor (equivalent to $k=0$) can reduce the total number of tunable parameters during fine-tuning,
%when applied to downstream datasets, 
but it leads to the severe performance degradation on all 6 datasets. This indicates the fixed model may impede sufficient adaption when dealing with various downstream datasets. 
By gradually increasing tunable layers $k$ ($1 \rightarrow 3$), the performance drop caused by insufficient adaption is mitigated. 
However, by tuning only partial model parameters, Last-$k$ still yield inferior results than the vanilla strategy (equivalent to $k=5$).
Adapter method, although has demonstrated competitive fine-tuning capacity with the vanilla strategy when fine-tuning language models in natural language processing domain,
%its competitive fine-tuning capacity with the standard strategy in other domains, 
fails to yield satisfactory results when fine-tuning GNNs for graph data.

To summarize, by investigating more fine-tuning strategies that are promising in other domains,
we conclude that they may not be suitable for fine-tuning GNNs, which is probably due to the domain divergence.
%divergence among different domains. 
%variant requirements of fine-tuning 
Instead,
specific and innovative designs to fine-tune GNNs in graph domain is much more demanded. 
The performance comparison
in Tab.~\ref{tab: exp_c2} also indicates that
the design dimensions in S2PGNN's search space
may be more validated for searching the suitable fine-tuning strategy in pre-trained GNNs.

\begin{table*}[!t]
	\caption{The performance comparison among S2PGNN's variants with degraded space.}
	\label{tab: exp_d}
	\centering
	\setlength\tabcolsep{7pt}
	%\scalebox{1}{
	\begin{tabular}{c|cccccc|cc|c}
		\hline
		\hline
		&
		\multicolumn{6}{c|}{\bf Classification (ROC-AUC (\%)) $\uparrow$}
		&\multicolumn{2}{c|}{\bf Regression (RMSE) $\downarrow$}
		&\multirow{2}{*}{\bf \makecell{Avg. \\ Drop}}
		\\
		\cline{1-9}
		\bf
		Dataset
		&BBBP 
		&Tox21 
		&ToxCast 
		&SIDER 
		&ClinTox
		&BACE 
		&ESOL 
		&Lipo
		&
		\\
		\hline
		S2PGNN-$\backslash id$
		&69.5 $\pm$ 2.0
		&75.9 $\pm$ 0.3
		&66.3 $\pm$ 0.4
		&61.3 $\pm$ 0.7
		&69.4 $\pm$ 6.3
		&79.5 $\pm$ 1.6
		&2.0 $\pm$ 0.3
		&0.9 $\pm$ 0.0
		& $-5.2\%$
		\\
		S2PGNN-$\backslash fuse$
		&69.0 $\pm$ 1.5
		&75.7 $\pm$ 0.6
		&65.7 $\pm$ 0.4
		&61.6 $\pm$ 0.7
		&61.6 $\pm$ 4.4
		&82.0 $\pm$ 1.0
		&2.5 $\pm$ 0.1
		&1.0 $\pm$ 0.0
		& $-12.1\%$
		\\
		S2PGNN-$\backslash read$
		&70.3 $\pm$ 1.6
		&75.2 $\pm$ 0.3
		&63.9 $\pm$ 0.3
		&62.2 $\pm$ 0.7
		&73.7 $\pm$ 4.4
		&80.3 $\pm$ 1.7
		&2.7 $\pm$ 0.0
		&1.0 $\pm$ 0.0
		& $-12.3\%$
		\\
		S2PGNN 
		&\textbf{70.9 $\pm$ 1.3} 
		&\textbf{76.3 $\pm$ 0.4} 
		&\textbf{67.0 $\pm$ 0.5} 
		&\textbf{62.8 $\pm$ 0.3} 
		&\textbf{75.9 $\pm$ 2.2} 
		&\textbf{82.6 $\pm$ 0.7} 
		&\textbf{1.7 $\pm$ 0.2}
		&\textbf{0.9 $\pm$ 0.0}
		&-
		\\
		\hline
		\hline
	\end{tabular}
	%	\vspace{-10px}
\end{table*}

\begin{table*}[!t]
	\caption{The performance comparison between proposed S2PGNN fine-tuning and vanilla fine-tuning strategies
		with fixed ContextPred pre-training objective and other popular GNN backbone architectures.}
	\label{tab: exp_b2}
	\centering
	%	\vspace{-10px}
	\setlength\tabcolsep{4pt}
	%\scalebox{1}{
	\begin{tabular}{c|cccccc|cc|c}
		\hline
		\hline
		&
		\multicolumn{6}{c|}{\bf Classification (ROC-AUC (\%)) $\uparrow$}
		&
		\multicolumn{2}{c|}{\bf Regression (RMSE) $\downarrow$}
		&\multirow{2}{*}{\bf \makecell{Avg. \\ Gain}}
		\\
		\cline{1-9}
		\bf 
		Dataset
		&BBBP 
		&Tox21 
		&ToxCast 
		&SIDER 
		&ClinTox
		&BACE 
		%		&\makecell{Avg. $\uparrow$}
		&ESOL 
		&Lipo
		&
		\\
		\hline
		ContextPred (GCN) 
		&64.6 $\pm$ 2.2
		&73.0 $\pm$ 0.5
		&61.9 $\pm$ 1.1
		&56.8 $\pm$ 0.6
		&69.0 $\pm$ 1.3
		&79.9 $\pm$ 1.7
		%		&
		&2.4 $\pm$ 0.1
		&\bf 1.0 $\pm$ 0.0
		&\multirow{2}{*}{$+4.6\%$}
		\\
		ContextPred (GCN) + S2PGNN
		&\bf 68.3 $\pm$ 1.0
		&\bf 75.7 $\pm$ 0.5
		&\bf 66.5 $\pm$ 0.3
		&\bf 62.3 $\pm$ 0.4
		&\bf 71.6 $\pm$ 1.3
		&\bf 81.5 $\pm$ 0.5
		%		&
		&\bf 2.3 $\pm$ 0.1
		&\bf 1.0 $\pm$ 0.0
		&
		\\
		\hline
		ContextPred (SAGE) 
		&65.0 $\pm$ 3.0
		&74.7 $\pm$ 0.5
		&63.4 $\pm$ 0.2
		&\bf 62.0 $\pm$ 0.6
		&61.1 $\pm$ 3.1
		&78.8 $\pm$ 1.4
		%		&
		&2.5 $\pm$ 0.1
		&\bf 1.0 $\pm$ 0.0
		&\multirow{2}{*}{$+6.0\%$}
		\\
		ContextPred (SAGE) + S2PGNN
		&\bf 69.0 $\pm$ 1.4
		&\bf 75.1 $\pm$ 0.4
		&\bf 66.4 $\pm$ 0.6
		&61.6 $\pm$ 0.4
		&\bf 67.1 $\pm$ 1.6
		&\bf 79.1 $\pm$ 0.7
		%		&
		&\bf 2.0 $\pm$ 0.2
		&\bf 1.0 $\pm$ 0.0
		&
		\\
		\hline
		ContextPred (GAT) 
		&64.9 $\pm$ 1.2
		&69.6 $\pm$ 0.7
		&59.5 $\pm$ 0.8
		&52.5 $\pm$ 3.4
		&58.2 $\pm$ 6.9
		&60.5 $\pm$ 3.5
		%		&
		&3.2 $\pm$ 0.1
		&1.1 $\pm$ 0.0
		&\multirow{2}{*}{$+19.7\%$}
		\\
		ContextPred (GAT) + S2PGNN
		&\bf 69.6 $\pm$ 0.9
		&\bf 75.0 $\pm$ 0.4
		&\bf 65.3 $\pm$ 0.3
		&\bf 61.8 $\pm$ 1.1
		&\bf 66.7 $\pm$ 3.2
		&\bf 80.4 $\pm$ 1.3
		%		&
		&\bf 1.8 $\pm$ 0.1
		&\bf 1.0 $\pm$ 0.0
		&
		\\
		\hline
		\hline
	\end{tabular}
	%	\vspace{-10px}
\end{table*}

\begin{table*}[!t]
	\caption{The running time (seconds per epoch) of several fine-tuning strategies.}
	\label{tab: exp_time}
	\centering
	%	\vspace{-10px}
	\setlength\tabcolsep{10pt}
	%\scalebox{1}{
	\begin{tabular}{c|cccccc|c}
		\hline \hline
		&
		\multicolumn{6}{c|}{\bf Classification}
		&\multirow{2}{*}{\bf Avg.}
		\\
		\cline{1-7}
		\bf Dataset
		&BBBP 
		&Tox21 
		&ToxCast 
		&SIDER 
		&ClinTox
		&BACE 
		&
		\\
		\hline

		Vanilla fine-tuning
		&5.2
		&14.0
		&11.3
		&3.3
		&2.7
		&6.8
		&7.2
		\\
		\hline
		\textit{$L^2$-SP}~\cite{xuhong2018explicit}
		&5.3
		&23.8
		&27.3
		&5.3
		&6.8
		&4.3
		&12.1
		\\
		DELTA~\cite{li2019delta}
		&5.7
		&11.4
		&11.1
		&5.8
		&3.2
		&5.0
		&7.0
		\\		
		BSS~\cite{chen2019catastrophic}
		&6.2
		&30.8
		&6.5
		&24.9
		&70.9
		&6.1
		&24.2
		\\
		StochNorm~\cite{kou2020stochastic}.
		&4.5
		&17.8
		&31.1
		&3.3
		&3.3
		&3.8
		&10.6
		\\
		\hline
		GTOT-tuning~\cite{zhang2022fine}
		&5.7
		&22.7
		&34.1
		&4.6
		&3.0
		&11.3
		&13.6
		\\
		\hline		
		S2PGNN 
		&13.3
		&16.5
		&18.3
		&13.3
		&18.0
		&14.0
		&15.6
		\\
		\hline
		\hline
	\end{tabular}
	%	\vspace{-10px}
\end{table*}

%{\color{red}
%\subsection{The rationality of S2PGNN's design space}
%\label{ssec: exp_d}
%
%
%this part can be discarded since we will discuss it in Sec. 4-C2
%
%}

\subsection{Ablation study on S2PGNN's design dimensions}
\label{ssec: exp_e}

To investigate S2PGNN's important design dimensions (see Sec. \ref{ssec: method_space}) regarding GNN fine-tuning strategy:
identity augmentation,
multi-scale fusion, 
and adaptive graph-level readout,
we further propose S2PGNN variants with degraded space and conduct ablation studies:
S2PGNN-$\backslash id$ disables identity augmentation when aggregating  messages from neighbors; 
S2PGNN-$\backslash fuse$ discards the multi-scale fusion and directly uses the last-layer output as learned node representations as most existing works does; 
S2PGNN-$\backslash read$ uses the simple and fixed mean pooling as \cite{hu2019strategies} and follow-up works.

The man results of S2PGNN's variants are shown in Tab.~\ref{tab: exp_d}.
Significant performances drop are observed in each S2PGNN variant with degraded space, which provide empirical validation that 
the proposed design dimensions in S2PGNN
are key factors that affect GNN fine-tuning results and should be incorporated during fine-tuning to achieve the optimal downstream results.

\subsection{Effect of GNN backbone architectures}
\label{ssec: exp_f}

Recent works \cite{sun2022does} have identified that GNN backbone is also crucial for GNN pre-training.
Therefore, here we further provide additional results of S2PGNN when built on top of other classic GNN backbone architectures other than the GIN.
Tab. \ref{tab: exp_b2}
summarizes the
performance of S2PGNN based on several pre-trained GNNs, including GCN, SAGE, and GAT models 
via the SSL strategy ContextPred \cite{hu2019strategies}.
We observe that pre-trained GNNs with all these backbone architectures can benefit from S2PGNN fine-tuning and achieve better performances than the vanilla strategy.
This verifies that S2PGNN is agnostic to the base GNN architectures and is capable to achieve the consistent improvement. 

%As shown in Table 3, we verify that Mole-BERT is agnostic to the
%GNN architectures by trying four popular GNN models including GIN (Xu et al., 2019), GCN (Kipf.
%& Welling, 2017), R-GCN (Schlichtkrull et al., 2018) and GraphSAGE (Hamilton et al., 2017). As
%can be observed, Mole-BERT achieves consistent and notable improvements over training from
%scratch with various GNNs. Additionally, pre-training with GIN achieves the most significant gains.

\subsection{Efficiency}
\label{ssec: exp_g}
Apart from the effectiveness results provided in previous subsections, here we further report the concrete running time comparisons among several fine-tuning baselines. 
As shown in Tab. \ref{tab: exp_time},
we observe that the running time of S2PGNN is comparable with fine-tuning baselines, which 
eliminates the concern that the fine-tuning search method in S2PGNN may consume more computing resources to achieve improved perform ances.
%{\color{blue}
As demonstrated in Remark \ref{remark: complexity} and Tab. \ref{tab: space},
given that the space complexity in S2PGNN is
$O(|\mathcal{O}_{conv}|^{K}\cdot |\mathcal{O}_{id}|^{K} \cdot |\mathcal{O}_{fuse}| \cdot |\mathcal{O}_{read}|)$ and the real space size built on the 5-layer GNN in our empirical study will have 10,206 candidate fine-tuning strategies,
it further verifies that the proposed search algorithm in Sec. \ref{ssec: method_alg} is indeed powerful to tackle the effiency challenge (as mentioned in Sec. \ref{sec: intro}) behind the fine-tuning search problem in this work.
%efficient for accelerating searching.
%}

%
%\begin{remark}[\color{blue} Space Complexity of $\bm{\Phi}_{ft}$]	
%	{\color{blue}
%		As shown in Fig.~\ref{fig: ours} and summarized in Tab. \ref{tab: space}, the overall complexity of the proposed GNN fine-tuning strategy space $\bm{\Phi}_{ft}$ in S2PGNN equals to the Cartesian product of the size of all involved dimensions, 
%		i.e., $O(|\mathcal{O}_{conv}|^{K}\cdot |\mathcal{O}_{id}|^{K} \cdot |\mathcal{O}_{fuse}| \cdot |\mathcal{O}_{read}|)$.
%		%$= 1^5 \times 3^5 \times 7 \times 6 =10206$.}
%	For the illustration in Fig.~\ref{fig: ours}, the search space built on the 5-layer GIN will have 10,206 candidate fine-tuning strategies.
%	Because each candidate will require to train the GNN parameter to convergence,
%	we cannot simply use a brute-force algorithm to test all possible candidates and train thousands of GNNs, which can lead to the enormous computational overhead.}
%\label{remark: complexity}
%\end{remark}

\section{Conclusion}
\label{sec: con}
%\footnote{\# zhili: may refer to BSS "Catastrophic Forgetting Meets Negative Transfer: Batch Spectral Shrinkage for Safe Transfer Learning"}
In this paper,
to 
fully unleash the potential of pre-trained GNNs on various downstream graph
and 
bridge the missing gap between better fine-tuning strategies with 
pre-trained GNNs,
we propose to search to fine-tune pre-trained 
%graph neural networks
GNNs
for 
graph-level tasks,
named S2PGNN.
%which can adaptively design a suitable fine-tuning framework for the given pre-trained GNN and downstream data.
%To improve the utilization of pre-trained GNNs,
%To fully unleash the potential of pre-trained GNNs on various downstream datasets, 
%S2PGNN adaptively designs a suitable fine-tuning strategy for the given pre-trained GNN
%and downstream dataset.
To achieve this, 
S2PGNN
first investigates fine-tuning within and outside GNN area to
%and provide a new perspective for GNN fine-tuning.
%present a novel search space for fine-tuning strategies,
identify key factors that affect GNN fine-tuning results
and carefully present a novel search space that is suitable for GNN fine-tuning.
To reduce the search cost from the large and discrete space,
we incorporate an efficient search algorithm to suggest the parameter-sharing and continuous relaxation 
%on the discrete space 
and solve the search problem by differentiable optimization. 
S2PGNN is model-agnostic and can be plugged into existing GNN backbone models
and pre-trained GNNs.
Empirical studies demonstrate that
S2PGNN can consistently improve 10 classic pre-trained GNNs
and achieve better performance than other fine-tuning works.
Therefore,
we expect S2PGNN to shed light on future directions towards better GNN fine-tuning innovation.
For the future works, it may be
worth trying to investigate how to 
derive a more robust and transferrable pre-trained GNNs 
that can be more easily adapted for various downstream graph scenarios.
%pre-train a more powerful GNN (e.g., from the prospective of adaptively searching a more expressive and transferable backbone architecture) that can be more easily transferred and fine-tuned to handle various downstream scenarios.

%{\color{red}
%+++
%for future works
%+++
%}

%\begin{table}[htbp]
%\caption{Table Type Styles}
%\begin{center}
%\begin{tabular}{|c|c|c|c|}
%\hline
%\textbf{Table}&\multicolumn{3}{|c|}{\textbf{Table Column Head}} \\
%\cline{2-4} 
%\textbf{Head} & \textbf{\textit{Table column subhead}}& \textbf{\textit{Subhead}}& \textbf{\textit{Subhead}} \\
%\hline
%copy& More table copy$^{\mathrm{a}}$& &  \\
%\hline
%\multicolumn{4}{l}{$^{\mathrm{a}}$Sample of a Table footnote.}
%\end{tabular}
%\label{tab1}
%\end{center}
%\end{table}

\section*{Acknowledgment}

%Lei Chen's work is partially supported by National Science Foundation of China (NSFC) under Grant No. U22B2060, the Hong Kong RGC GRF Project 16209519, CRF Project C6030-18G, C2004-21GF, AOE Project AoE/E-603/18, RIF Project R6020-19, Theme-based project TRS T41-603/20R, China NSFC No. 61729201, Guangdong Basic and Applied Basic Research Foundation 2019B151530001, Hong Kong ITC ITF grants MHX/078/21 and PRP/004/22FX, Microsoft Research Asia Collaborative Research Grant, HKUST-Webank joint research lab grant and HKUST Global Strategic Partnership Fund (2021 SJTU-HKUST). 
Lei Chen's work is partially supported by National Science Foundation of China (NSFC) under Grant No. U22B2060, the Hong Kong RGC GRF Project 16213620, RIF Project R6020-19, AOE Project AoE/E-603/18, Theme-based project TRS T41-603/20R, CRF Project C2004-21G, China NSFC No. 61729201, Guangdong Basic and Applied Basic Research Foundation 2019B151530001, Hong Kong ITC ITF grants MHX/078/21 and PRP/004/22FX, Microsoft Research Asia Collaborative Research Grant and HKUST-Webank joint research lab grants.
Xiaofang Zhou's work is supported by the JC STEM Lab of Data Science Foundations funded by The Hong Kong Jockey Club Charities Trust, HKUST-China Unicom Joint Lab on Smart Society, and HKUST-HKPC Joint Lab on Industrial AI and Robotics Research.

\clearpage
\newpage

\bibliographystyle{ieeetr}
\bibliography{sample-base}

\end{document}